\pdfoutput=1

\documentclass[11pt]{article}

\usepackage[final]{acl}

\usepackage{times}
\usepackage{latexsym}

\usepackage[T1]{fontenc}

\usepackage[utf8]{inputenc}

\usepackage{microtype}

\usepackage{inconsolata}

\usepackage{graphicx}

\usepackage{CJKutf8}

\usepackage{hyperref}
\usepackage{url}

\usepackage{amssymb}
\usepackage{amsmath}
\usepackage{amsthm}
\usepackage{amsfonts}
\usepackage{bm}
\usepackage{nicefrac}
\usepackage{dsfont}

\usepackage{booktabs}
\usepackage{multirow}
\usepackage{makecell}
\usepackage{arydshln}

\usepackage{color}
\usepackage{xcolor}
\usepackage{colortbl}

\usepackage{comment}
\usepackage{xspace}
\usepackage{soul}

\usepackage{enumitem}
\usepackage[normalem]{ulem}
\setlist[itemize,enumerate]{leftmargin=*}

\makeatletter
\def\adl@drawiv#1#2#3{%
        \hskip.5\tabcolsep
        \xleaders#3{#2.5\@tempdimb #1{1}#2.5\@tempdimb}%
                #2\z@ plus1fil minus1fil\relax
        \hskip.5\tabcolsep}
\newcommand{\cdashlinelr}[1]{%
  \noalign{\vskip 2pt
           \global\let\@dashdrawstore\adl@draw
           \global\let\adl@draw\adl@drawiv}
  \cdashline{#1}[.4pt/2pt]
  \noalign{\global\let\adl@draw\@dashdrawstore
           \vskip 2pt}}
\makeatother

\renewenvironment{quote}{%
  \list{}{%
    \leftmargin0.5cm   %
    \rightmargin\leftmargin
  }
  \item\relax
}
{\endlist}

\definecolor{light-orange}{HTML}{fee9d4}
\definecolor{light-green}{HTML}{d8f0d3}
\definecolor{light-blue}{HTML}{dae8f5}
\definecolor{light-red}{HTML}{FBC7C4}
\definecolor{set10-red}{HTML}{e41a1c}
\definecolor{set10-blue}{HTML}{377eb8}
\definecolor{set10-green}{HTML}{4daf4a}

\definecolor{CustomBlue}{RGB}{57,83,191}
\usepackage[most]{tcolorbox}

\newtcbox{\clustertab}[1]{on line, box align=base, colback={#1},colframe={#1},size=fbox,arc=2pt,top=-1.5pt, bottom=-1.5pt, left=-1.5pt, right=-1.5pt, boxrule=0pt, enlarge left by=1pt}
\newcommand{\firstcluster}[1]{{\tiny\clustertab{set10-blue!40}{#1}}}
\newcommand{\secondcluster}[1]{{\tiny\clustertab{set10-blue!25}{#1}}}
\newcommand{\thirdcluster}[1]{{\tiny\clustertab{set10-blue!20}{#1}}}
\newcommand{\fourthcluster}[1]{{\tiny\clustertab{set10-blue!10}{#1}}}

\newcommand{\secondbadcluster}[1]{{\tiny\clustertab{set10-red!25}{#1}}}
\newcommand{\thirdbadcluster}[1]{{\tiny\clustertab{set10-red!20}{#1}}}
\newcommand{\fourthbadcluster}[1]{{\tiny\clustertab{set10-red!10}{#1}}}
\newcommand{\fifthbadcluster}[1]{{\tiny\clustertab{set10-red!5}{#1}}}
\newcommand{\othercluster}[1]{{\tiny\clustertab{white!0}{\phantom{#1}}}}

\newcommand{\Tower}{\textsc{Tower}\xspace}
\newcommand{\xTower}{\textsc{xTower}\xspace}
\newcommand{\TowerBase}{\textsc{TowerBase}\xspace}
\newcommand{\TowerInstruct}{\textsc{TowerInstruct}\xspace}
\newcommand{\Comet}{\textsc{Comet}\xspace}
\newcommand{\xComet}{\textsc{xComet}\xspace}

\title{\xTower:  Machine Translation with Fine-grained Error Explanations}
\title{\xTower: \\ A Multilingual LLM for Explaining and Correcting Translation Errors}

\author{
 \textbf{Marcos Treviso\textsuperscript{1}},
 \textbf{Nuno M. Guerreiro\textsuperscript{1,2,3,4}},
 \textbf{Sweta Agrawal\textsuperscript{1}},
 \textbf{Ricardo Rei\textsuperscript{3}},
 \textbf{José Pombal\textsuperscript{3}},
 \\
 \textbf{Tania Vaz\textsuperscript{3}},
 \textbf{Helena Wu\textsuperscript{3}},
 \textbf{Beatriz Silva\textsuperscript{3}},
 \textbf{Daan van Stigt\textsuperscript{3}},
 \textbf{André F. T. Martins\textsuperscript{1,2,3,5}}
\\
 \normalsize \textsuperscript{1}Instituto de Telecomunicações, \textsuperscript{2}Instituto Superior Técnico, Universidade de Lisboa, \\
 \normalsize \textsuperscript{3}Unbabel,
 \textsuperscript{4}MICS, CentraleSupélec, Université Paris-Saclay,
 \textsuperscript{5}ELLIS Unit Lisbon
}

\begin{document}
\maketitle
\begin{abstract}
While machine translation (MT) systems are achieving increasingly strong performance on  benchmarks, they often produce translations with errors and anomalies. 
Understanding these errors can potentially help improve the translation quality and user experience.
This paper introduces \xTower, 
an open large language model (LLM) built on top of \TowerBase designed to provide free-text explanations for translation errors in order to guide the generation of a corrected translation.
The quality of the generated explanations by \xTower are assessed via both intrinsic and extrinsic evaluation.
We ask expert translators to evaluate the quality of the explanations across two dimensions: \emph{relatedness} towards the error span being explained and \emph{helpfulness} in error understanding and improving translation quality. 
Extrinsically, we test \xTower across various experimental setups in generating translation corrections, demonstrating significant improvements in translation quality.  Our findings highlight \xTower's potential towards not only producing plausible and helpful explanations of automatic translations, but also leveraging them to suggest corrected translations.\footnote{\url{http://huggingface.co/sardinelab/xTower13B}}
\end{abstract}

\section{Introduction}

Neural machine translation (MT) systems have made significant strides in recent years. 
However, despite their high performance on standard benchmarks, these systems often produce translations that contain errors and anomalies. 
Common methods for evaluating MT quality, such as BLEU~\citep{papineni-etal-2002-bleu}, and neural metrics like \Comet~\citep{rei-etal-2020-comet} and BLEURT \cite{sellam-etal-2020-bleurt}, provide only a numerical score reflecting overall translation quality. 
Recent metrics like \xComet~\citep{guerreiro2023xcomet} and \textsc{AutoMQM}~\citep{fernandes-etal-2023-devil} highlight error spans to justify their scores but do not offer explanations about the nature of these errors.
InstructScore, a recent work by \citet{xu-etal-2023-instructscore}, leverages large language models (LLMs) to provide a quality score conditioned on built-in error detection and explanations.
However, InstructScore primarily functions as a \emph{reference-based metric},  using explanations as a means to improve score estimates via meta-feedback/finetuning.

\begin{figure*}[t]
    \centering
    \includegraphics[width=\textwidth]{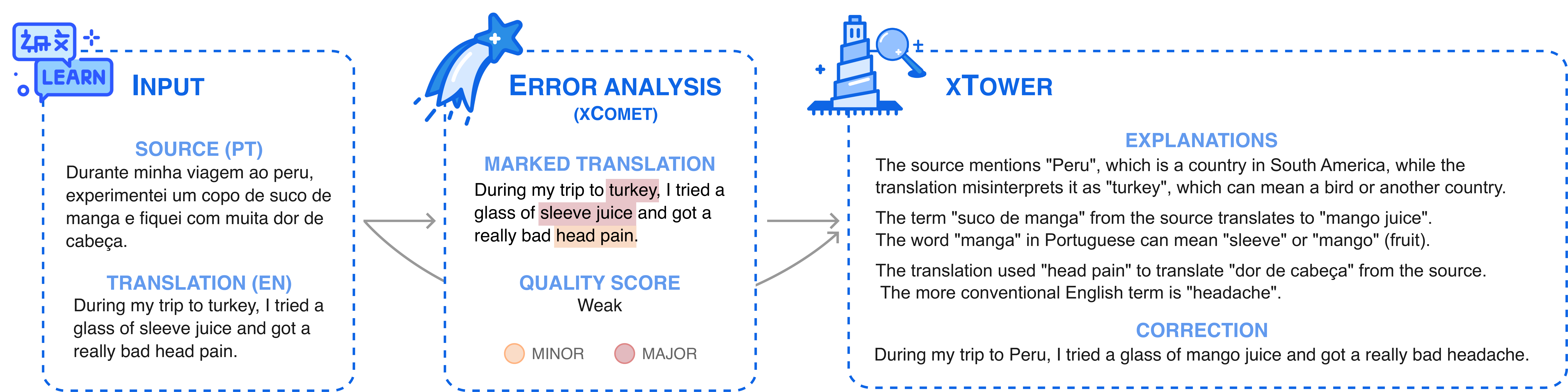}
    \caption{Illustration of our approach. In this example, the input consisting of a source and a translation is passed to 
    \xComet, which annotates the translation with error spans and produces a (discretized) quality score. The full input, marked translation, and quality score are passed to \xTower, which, in turn, produces an explanation for each error span along with a final suggestion for a new, corrected translation.}
    \label{fig:xTower}
\end{figure*}

In this paper, we introduce \xTower (Figure~\ref{fig:xTower}), a LLM specifically tailored to produce high-quality explanations for translation errors and to utilize these explanations to suggest corrections through chain-of-thought prompting~\citep{wei2023chainofthought}.
\xTower is built on \TowerBase 13B~\citep{alves2024tower}, a strong open multilingual LLM for MT-related tasks. 
Unlike InstructScore, 
\xTower can operate \emph{without the need for reference translations} while also considering information contained in the source sentence. 
Moreover, \xTower is designed to be agnostic about the source of error spans, as they can be obtained manually via human annotation or via automatic tools. 
In this work, we experiment with both.
For the automatic case, we leverage \xComet~\citep{guerreiro2023xcomet}. 
This modular approach offers flexibility to experiment with span-level error annotations from various sources, and easily incorporate future improvements in span error detection tools without requiring retraining.

We evaluate \xTower's explanations both intrinsically and extrinsically. 
Intrinsically, we employ human evaluation to score explanations on two dimensions: \emph{relatedness} to the error spans being explained (\S\ref{sec:relatedness}) and \emph{helpfulness} in guiding towards a better translation (\S\ref{sec:helpfulness}). 
Extrinsically, we assess \xTower's ability to suggest translation corrections (\S\ref{sec:refining_translations}), experimenting with different error span sources (human vs. predicted). 
We compare \xTower's performance against leading closed and open LLMs, such as GPT-3.5 Turbo, Mixtral 8x7B, and \TowerInstruct 13B. 
Our findings demonstrate that \xTower improves error interpretability by providing explanations that effectively relate to the marked errors. 
Expert translators endorse these explanations as helpful for understanding translation errors and generally useful for improving translations, particularly for English-German. 
Furthermore, prompting \xTower with these explanations  leads to improved translation corrections.
Overall, our main contributions are:

\begin{itemize}
    \item We introduce \xTower, a multilingual LLM that generates free-text explanations for translation errors and provides corrected translations.
    
    \item We conduct extensive human evaluations to assess the relatedness and helpfulness of \xTower's explanations, linking their results with dedicated qualitative analyses.

    \item We evaluate \xTower's corrected translations across multiple language pairs and experimental setups, showing significant improvements in translation quality.
    
\end{itemize}

\section{Background}

In this section, we provide an overview of the key components and concepts relevant to our work. %

\paragraph{\Tower.} \citet{alves2024tower} developed a suite of state-of-the-art multilingual LLMs via continued pretraining of LLaMA2 \citep{touvron2023llama} --- \TowerBase ---~and supervised finetuning for translation-related tasks --- \TowerInstruct.
\Tower is trained to handle diverse tasks such as MT, automatic post-editing, and grammatical error correction. 
However, it lacks support for error-annotated inputs and cannot produce high-quality, span-level explanations for translation errors. 
\xTower addresses these limitations by extending \Tower --- through distillation and finetuning --- enabling it to provide explanations for translation errors and generate corrected translations.

\paragraph{MT Evaluation.} Evaluating the quality of machine translations is a critical aspect of improving MT systems. Traditional metrics like BLEU~\citep{papineni-etal-2002-bleu} and \textsc{chrF}~\citep{popovic-2015-chrf} have been widely used to measure the accuracy of translations by comparing them to reference translations. However, these lexical metrics do not correlate well with human judgments~\cite{freitag-etal-2023-results}. 
More recent neural metrics, such as BLEURT~\citep{sellam-etal-2020-bleurt} and \Comet~\citep{rei-etal-2020-comet}, offer improved performance by finetuning pretrained neural models to predict translation quality. 
Still, they lack the ability to explain errors in human-interpretable terms. 
To this end, \citet{rei-etal-2023-inside,guerreiro2023xcomet} propose methods to highlight input words relevant to the output.
However, highlighting input words offers a limited view of interpretability, as the end-user often needs additional information to understand what the error consists of and how it can be fixed.
Our approach with \xTower aims to bridge this gap by generating free-text explanations for translation errors, thus offering more insightful and detailed quality reports.

\section{\xTower}

In this section, we provide details on the methodology behind \xTower (Figure~\ref{fig:xTower}), a model built on top of \TowerBase via distilled supervised finetuning~\citep{tunstall2023zephyr}.

\subsection{Distillation} 
\label{subsec:distillation}

\paragraph{Data.} We use GPT-4 to generate explanations 
for samples annotated with MQM spans and to generate a final translation correction.\footnote{We use \texttt{gpt-4-0125} available from the OpenAI API.} 
Our dataset comprises English$\to$German (\textsc{en-de}), English$\to$Russian (\textsc{en-ru}), and Chinese$\to$English (\textsc{zh-en}) samples from the WMT 2022 Metric shared task~\citep{freitag-etal-2022-results}.
Each error span is annotated by humans according to the MQM framework, which includes a severity rating such as minor or major.
Detailed statistics about this dataset are provided in  Appendix~\ref{sec:datasets_statistics}.
Overall, our distillation dataset consists of 33,442 samples containing 63,188 human-annotated error spans.

\paragraph{Prompt.} We use an XML format to obtain an ``annotated translation'', which includes the demarcations of error spans as tags alongside their severity as attributes. 
Following \citet{farinha-etal-2022-findings}, we discretize the MQM quality score into buckets: weak, moderate, good, excellent, best. 
Table~\ref{tab:example_prompt_gpt4} shows a prompt example.
As output, GPT-4 generates explanations for each marked error, followed by a corrected translation in the following format:%
\footnote{We manually inspected a few outputs to ensure reliability.} 
\begin{itemize}
    \item \texttt{Explanation for error$N$}: the explanation given to the $N\textsuperscript{th}$ error span. Explanations for each error span are separated by newlines.
    \item \texttt{Translation correction}: the translation refinement produced by the model. Corrections are placed in a new line after the last explanation.
\end{itemize} 

Notably, this ordering acts as chain-of-thought prompting~\citep{wei2023chainofthought}. 
We collect outputs for referenceless and reference-based evaluation~(by providing a reference translation in the input).

\begin{table}[t]
    \small
    \centering
    \begin{tabular}{@{}m{0.999\columnwidth}@{}}
    \toprule

        \textcolor{gray}{\textit{Instruction:}} \\
        You are provided with a Source, Translation, Translation quality analysis, and Translation quality score (weak, moderate, good, excellent, best). The Translation quality analysis contains a translation with marked error spans with different levels of severity (minor or major). Given this information, generate an explanation for each error and a fully correct translation. \\
        \cdashlinelr{1-1}
        \textcolor{gray}{\textit{Input:}} \\
        \textbf{English source:} All were wearing avalanche beacons. \\
        \textbf{German translation:} Alle trugen Lawinenschilder. \\
        \textbf{Translation quality analysis:}
        Alle trugen 
        <error1 severity=``major''>Lawinenschilder</error1>.
        \\
        \textbf{Translation quality score:} weak \\
        \cdashlinelr{1-1}
        \textcolor{gray}{\textit{Output:}}
        \\
        \textbf{Explanation for error1:} The term ``Lawinenschilder'' translates to ``avalanche signs'', which is a significant mistranslation of ``avalanche beacons''. Avalanche beacons are specific devices used to locate people trapped under snow after an avalanche, not signs. This error changes the meaning of the sentence and introduces confusion about the context. \\
        \textbf{Translation correction:} Alle trugen Lawinensuchgeräte. \\

    \bottomrule
    \end{tabular}
    \caption{Example of prompt used for generating explanations and translation corrections.}
    \label{tab:example_prompt_gpt4}
\end{table}

\subsection{Finetuning} 
\label{subsec:finetuning}

We obtained \textsc{xTower} by finetuning \textsc{TowerBase-13b} on a dataset that includes the GPT-4 generated explanations described in \S\ref{subsec:distillation}, and machine translation data from \textsc{TowerBlocks}, the dataset used to train \textsc{TowerInstruct}.\footnote{\url{https://huggingface.co/datasets/Unbabel/TowerBlocks-v0.1}} We combined all available data to train a single, multilingual model, instead of training separate models for each language pair.
Moreover, following~\citep{longpre2023flan}, we employed a mixed prompt setting~(zero-shot, few-shot) during training. 
As a result, \textsc{xTower} can handle both referenceless and reference-based $k$-shot prompts.
Our training hyperparameters and configuration follows that used to train \textsc{TowerInstruct}~\citep{alves2024tower}.

\section{Explaining Translation Errors}
\label{sec:explaining_translation_errors}

In this section, we provide a detailed human evaluation of the quality of the explanations produced by \xTower, which are obtained in a more realistic setting via referenceless prompting.

\subsection{Experimental Setup}

\paragraph{Data.} We evaluate our models on MQM annotations from the WMT 2023 Metrics shared task test set~\citep{kocmi-etal-2023-findings}, spanning three language pairs: English$\to$German (\textsc{en-de}), Hebrew$\to$English (\textsc{he-en}), and Chinese$\to$English (\textsc{zh-en}).
This dataset contains 24,781 samples with 69,564 human-annotated error spans.
To obtain a fully automatic approach, we use error spans predicted by \xComet-XL~\citep{guerreiro2023xcomet}.\footnote{\url{https://huggingface.co/Unbabel/XCOMET-XL}}
For a consistent evaluation, we also query \xComet without references for our experiments. 
In total, we obtain a set of 108,507 spans, indicating that \xComet has a higher tendency to predict errors.
Detailed statistics are shown in Appendix~\ref{sec:datasets_statistics}.

\paragraph{Prompting.} 
We use the same prompt template as the one used in our distillation experiments, 
shown in Table~\ref{tab:example_prompt_gpt4}. We use 0-shot prompting for all experiments involving \xTower in this section.

\paragraph{Evaluation.}

While recent works propose frameworks to assess free-text explanations for classification tasks~\citep{wiegreffe-etal-2021-measuring,ramnath2023tailoring,joshi-etal-2023-machine,chen2023models}, applying a similar evaluation for MT is challenging due to the occurrence of multiple error spans with varied impact on translation quality.
Therefore, we choose to assess our explanations through human evaluation and qualitative analysis (\S\ref{sec:qualitative}). 
The evaluation comprises the following two dimensions:
\begin{itemize}

    \item \textbf{Relatedness}: The extent to which the explanation is related to the content of the error span.
    
    \item \textbf{Helpfulness}: The extent to which the explanation helps in understanding the nature of the error and in guiding towards a translation correction.

\end{itemize}

We present the setup and findings from both evaluations next.
Human evaluation details and guidelines can be found in Appendix~\ref{sec:human_annotation}.

\begin{table}[t]
    \small
    \centering
    \setlength{\tabcolsep}{3.5pt}
    \begin{tabular}{ll@{} cc c@{} cc}
        \toprule
        & & \multicolumn{2}{c}{\sc en-de} 
        & & \multicolumn{2}{c}{\sc zh-en} \\ 
        \cmidrule{3-4} \cmidrule{6-7}
        \sc Level & & 
        \sc xComet & \sc Human & &
        \sc xComet & \sc Human \\
        \midrule
        
        Explanation & & 
            3.5\textcolor{gray}{\scriptsize{$\pm$1.5}} & 4.4\textcolor{gray}{\scriptsize{$\pm$1.6}} & &
            3.4\textcolor{gray}{\scriptsize{$\pm$1.6}} & 4.3\textcolor{gray}{\scriptsize{$\pm$1.7}} \\

        Document & &
            3.4\textcolor{gray}{\scriptsize{$\pm$1.5}} & 4.3\textcolor{gray}{\scriptsize{$\pm$1.7}} & &
            3.3\textcolor{gray}{\scriptsize{$\pm$1.6}} & 4.3\textcolor{gray}{\scriptsize{$\pm$1.7}} \\

        \cdashlinelr{1-7}
        Correlation & & 0.96 & 0.89 & & 0.96 & 0.96 \\
    \bottomrule
    \end{tabular}
    \caption{Relatedness scores (6-Likert scale) computed at explanation and document-level, along with the Spearman correlation between the two.}
    \label{tab:relatedness_results}
\end{table}

\subsection{Relatedness}
\label{sec:relatedness}

A total of 6 annotators were employed for the task, evaluating samples marked with \xComet and human-annotated error spans. 
3 annotators assessed explanations for \textsc{en-de}, and other 3 for \textsc{zh-en}. 
For each language pair and error span source, we randomly sampled 50 translations, resulting in 200 examples in total. 
Inspired by the direct assessment and scalar
quality metric (DA+SQM) scale used in MT evaluation~\citep{kocmi-etal-2022-findings},
we asked annotators to rate explanations on a 6-point Likert scale: nonsense/unrelated (0), somewhat (2), mostly (4), and fully related (6).\footnote{The full scoring rubric is provided in Appendix~\ref{sec:human_annotation}.}
Moreover, we asked annotators to rate the quality of explanations individually (\textbf{explanation-level}) and by looking at all explanations at once (\textbf{document-level}).
The annotations were carried out on the Upwork platform.\footnote{\url{https://www.upwork.com}}
We obtain an overall inter-annotator agreement, as measured via Spearman correlation \citep{pavlick-tetreault-2016-empirical}, of 0.51 (\textsc{en-de}) and 0.40 (\textsc{zh-en}) at the explanation-level, and of 0.50 (\textsc{en-de}) and 0.37 (\textsc{zh-en}) at the document-level,
suggesting a fair-to-moderate agreement among annotators, typical in explanations evaluation which is a subjective task~\citep{wiegreffe-etal-2022-reframing, kunz-etal-2022-human}. 
Results are shown in Table \ref{tab:relatedness_results}.

\begin{figure}[t]
    \centering
    \includegraphics[width=0.9\columnwidth]{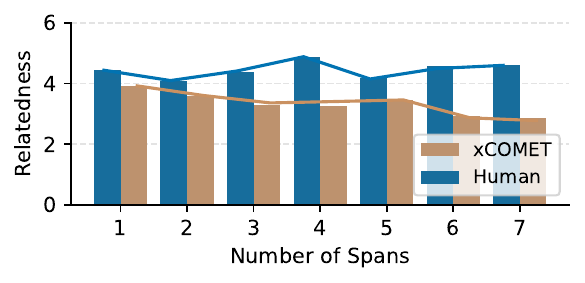}
    \vspace*{-0.25cm} 
    \caption{Relatedness according to the number of spans for \xComet and human error spans.}
    \label{fig:relatedness_scores_per_span_size}
\end{figure}

\paragraph{Discussion.} For human-annotated error spans, the overall relatedness scores range around 4.3, while for \xComet spans the scores drop to around 3.2. 
This difference indicates that \textbf{the quality of error spans heavily impacts the quality of their explanations}.
Nonetheless, for both cases, human ratings are in the 3-5 range, indicating that \textbf{\xTower's explanations are \textit{mostly} related to the error spans}. 
We also note a very high correlation between the quality of explanations assessed at the explanation and document-level, especially for human-annotated spans, 
indicating that the quality of explanations is consistent across granularities.

In Figure~\ref{fig:relatedness_scores_per_span_size} we show how relatedness scores vary according to the number of error spans.
We observe that, while the number of spans does not affect the relatedness of explanations produced for human-annotated error spans, they lead to a slight decrease of the relatedness scores when the spans are predicted by \xComet. We hypothesize this is due to \xComet overpredicting error spans (see Table~\ref{tab:data_stats}).

\subsection{Helpfulness}
\label{sec:helpfulness}

To quantify the idea of how helpful explanations are to the end user, we carried a new human evaluation with 4 of the same annotators from the previous task, and asked them to rate explanations based on two questions:
\begin{itemize}
    \item Q1: How helpful is the explanation in improving the understanding of the nature of the error?
    \item Q2: How helpful is the explanation in guiding towards writing a better translation?
\end{itemize} 
The rating is again performed on a 6-point Likert scale, ranging from less to more helpful. 
Moreover, we focus on studying the helpfulness of correct error spans only, in order to isolate the effect of providing accurate information towards improving error understanding.
To this end, we filter out samples with an overall relatedness score lower than 4 and only use error spans labeled by humans.
Table~\ref{tab:helpfulness_results} shows the results.

\paragraph{Discussion.} We find that annotators mark the explanations as being on average helpful (scores range in 4.4-4.6) in improving error understanding for both language pairs. Here, scores over 4 imply that ``the explanation clearly identifies the error and provides relevant details about its nature''. Furthermore, the usefulness of these explanations in guiding towards a potential correction ranges on average between 3.3-3.9, demonstrating that the explanations do hint towards a potential solution for correction, but they can be made more specific. For example, one of our expert annotators quoted:
\begin{quote}
    \small{
    Many cases had a very clear explanation of the nature of the error, but in terms of helpfulness in guiding towards writing a correction, it was a bit less clear than the above-mentioned examples as they do not suggest a correction. Nonetheless, the explanation still correctly guides the editor to a post-edition.
    }
\end{quote}

\begin{table}[t]
    \small
    \centering
    \begin{tabular}{lcc}
        \toprule 
        \sc Question & \sc en-de & \sc zh-en \\ 
        \midrule
        Q1: error understanding & 
            4.6\textcolor{gray}{\scriptsize{$\pm$1.7}} & 4.4\textcolor{gray}{\scriptsize{$\pm$2.1}} \\

        Q2: translation guidance &
            3.9\textcolor{gray}{\scriptsize{$\pm$1.6}} & 3.3\textcolor{gray}{\scriptsize{$\pm$2.1}} \\
        
        \cdashlinelr{1-3}
        Correlation & 0.85 & 0.72 \\

    \bottomrule
    \end{tabular}
    \caption{Helpfulness scores (6-Likert scale) for Q1 (understanding the nature of the errors) and Q2 (guiding towards better translations), along with the Spearman correlation between the two.}
    \label{tab:helpfulness_results}
\end{table}

\begin{table*}[t]
    \scriptsize
    \centering
    \setlength{\tabcolsep}{4.5pt}

    \begin{tabular}{ll}
    \toprule
    \sc Category & \sc Example \\
    \midrule
    \begin{tabular}[l]{@{}m{0.16\textwidth}@{}}
    \bf Correct Span with \\ 
    \bf Accurate Explanation \\
    \ \\
    Prevalence: 31.3\% \\
    Relatedness: 5.3
    \end{tabular} 
    & 
    \begin{tabular}[l]{@{}m{0.78\textwidth}@{}}
    \textbf{Source:} Indian govt withdraws appeal to hug cows on Valentine's Day after criticism
    \\
    \textbf{Reference:} Indische Regierung zieht Aufruf zur Umarmung von Kühen am Valentinstag nach Kritik zurück
    \\
    \textbf{Translation:} \clustertab{light-red}{\textbf{Indian Govt}} zieht Berufung zurück, Kühe am Valentinstag nach Kritik zu umarmen
    \\
    \textbf{Explanation:} The term "Indian Govt" is a direct transliteration of "Indian govt" from English to German, which is grammatically incorrect in German. The correct form should use the German spelling for "government" to match German grammatical rules, which would be "Indische Regierung."
    \end{tabular}
    \\
    \cdashlinelr{1-2}
    \begin{tabular}[l]{@{}m{0.16\textwidth}@{}}
    \bf Correct Span with \\ 
    \bf Inaccurate Explanation \\
    \ \\
    Prevalence: 14.1\% \\
    Relatedness: 1.6
    \end{tabular} 
    & 
    \begin{tabular}[l]{@{}m{0.78\textwidth}@{}}
    \textbf{Source:} Is it so ephemeral then? Merely a "catchphrase" with no clear originator?
    \\
    \textbf{Reference:} Ist es also so vergänglich? Lediglich ein „Sprichwort“ ohne eindeutigen Urheber?
    \\
    \textbf{Translation:} 
    Ist er also so kurzlebig? Lediglich ein \clustertab{light-red}{\textbf{"}}Schlagwort" ohne klaren Urheber?
    \\
    \textbf{Explanation:} The quotation marks around "Schlagwort" are correctly placed, but the analysis might be pointing out a minor stylistic preference or a typographical inconsistency. In German, the use of quotation marks can vary, but the provided translation is not incorrect in this regard. \\
    \end{tabular}
    \\
    \cdashlinelr{1-2}
    \begin{tabular}[l]{@{}m{0.16\textwidth}@{}}
    \bf Incorrect Span with \\ 
    \bf Valuable Explanation \\
    \ \\
    Prevalence: 27.8\% \\
    Relatedness: 5.3
    \end{tabular} 
    & 
    \begin{tabular}[l]{@{}m{0.78\textwidth}@{}}
    \textbf{Source:} To agonize over personal failure \& guilt in a world where "success" is so often bestowed on people responsible for the most reprehensible acts...
    \\
    \textbf{Reference:} Sich in einer Welt, in der „Erfolg“ so oft Menschen zuteilwird, die für die schlimmsten Taten verantwortlich sind, mit persönlichem Versagen und Schuldgefühlen herumzuquälen ... 
    \\
    \textbf{Translation:} Sich über persönliches Versagen und \clustertab{light-red}{\textbf{Schuld zu quälen}} in einer Welt, in der "Erfolg" so oft Menschen zuteilwird, die für die verwerflichsten Taten verantwortlich sind...
    \\
    \textbf{Explanation:} The phrase "Schuld zu quälen" is a direct translation of "agonize over guilt." The error marking might suggest a preference for a different phrasing or structure to convey the depth of emotional distress. However, "Schuld zu quälen" is a correct and idiomatic way to express the idea of being tormented by guilt in German.
    \end{tabular}
    \\
    \cdashlinelr{1-2}
    \begin{tabular}[l]{@{}m{0.16\textwidth}@{}}
    \bf Incorrect Span with \\ 
    \bf Worthless Explanation \\
    \ \\
    Prevalence: 26.83\% \\
    Relatedness: 1.3 \\ 
    \end{tabular} 
    & 
    \begin{tabular}[l]{@{}m{0.78\textwidth}@{}}
    \textbf{Source:}  Balenciaga boss calls holiday campaign 'stupid mistake'
    \\
    \textbf{Reference:} Balenciaga-Chef nennt Feiertagskampagne ‚dummen Fehler‘
    \\
    \textbf{Translation:} Balenciaga\clustertab{light-red}{\textbf{-Chef}} nennt Weihnachtskampagne „dummer Fehler“
    \\
    \textbf{Explanation:} The use of a hyphen between "Balenciaga" and "Chef" is not incorrect but could be seen as stylistically awkward in German. 
    The hyphen here might make the name of the position seem less formal or less integrated into the sentence. \\
    \end{tabular}
    \\
    \bottomrule
    \end{tabular}
    \caption{Examples of explanations according to our qualitative analysis. Error spans are highlighted in \colorbox{light-red}{red}.
    }
    \label{tab:qualitative_analysis}
\end{table*}

\subsection{Qualitative Analysis}
\label{sec:qualitative}

Based on the annotators' feedback for the previous experiments, and by manually inspecting the annotated examples, we present a qualitative analysis of the explanations generated by \xTower in Table \ref{tab:qualitative_analysis}.
Our analysis reveals several interesting scenarios that highlight \xTower's strengths and weaknesses. We categorize our findings into four main groups:

\begin{itemize}
    \item \textbf{Correct Spans:} 
    For error spans that correctly correspond to an error in the translation, explanations are \textbf{accurate} when they effectively detail the nature of the error, and \textbf{inaccurate} when they are unattached to the error, possibly suggesting wrong modifications.  

    \item \textbf{Incorrect Spans}: 
    Despite incorrect spans, explanations can still be \textbf{valuable} by pointing out that there are no errors in the translation. 
    In other cases, they are mislead by the incorrect span and become \textbf{worthless} by being nonsensical to the error, possibly including a boilerplate suggestion for stylistic improvement.

\end{itemize}

We also estimate the prevalence and compute the average relatedness score of each category. 
Specifically, 
we consider explanations as accurate/valuable when their average relatedness score is larger or equal to 4, otherwise we consider them as inaccurate/worthless.
This analysis indicates that \xTower is not only capable of generating high-quality explanations when the error spans are correctly identified, but can also provide valuable explanations for incorrect spans, amounting to 59.1\% of the cases with an average relatedness score of 5.3. 
However, over a quarter of all explanations (26.8\%) either misidentify the nature of the error or provide generic, boilerplate suggestions. 
These findings suggest that while \xTower has the potential to be a useful tool for automatic translation error analysis, there is still significant room for improvement, especially for cases where translation errors spans are incorrectly identified.

\begin{table*}[t]
    \small
    \centering
    \setlength{\tabcolsep}{3pt}
    \begin{tabular}{l lll c@{\ \ } lll c@{\ \ } lll}
        \toprule
        & \multicolumn{3}{c}{\sc en-de} & & 
        \multicolumn{3}{c}{\sc he-en} & & 
        \multicolumn{3}{c}{\sc zh-en} \\ 
        \cmidrule{2-4} \cmidrule{6-8} \cmidrule{10-12} 
        \sc Model & 
        \sc bleurt & \sc comet & \sc ckiwi & &
        \sc bleurt & \sc comet & \sc ckiwi & &
        \sc bleurt & \sc comet & \sc ckiwi \\
        \midrule

        Original MT & 48.4 \othercluster{0.0} & 78.4 \othercluster{0.0} & 75.5 \othercluster{0.0} & & 59.8 \othercluster{0.0} & 77.5 \othercluster{0.0} & 75.5 \othercluster{0.0} & & 55.2 \othercluster{0.0} & 78.0 \othercluster{0.0} & 76.7 \othercluster{0.0} \\

        \cdashlinelr{1-12}
        \multicolumn{10}{l}{\textcolor{gray}{\textit{Translation-only LLMs:}}} \\

        Mixtral 8x7B & 46.4 \fourthbadcluster{$\downarrow$ 2.0} & 80.4 \thirdcluster{$\uparrow$ 2.0} & 76.6 \thirdcluster{$\uparrow$ 1.1} & & 53.9 \secondbadcluster{$\downarrow$ 5.9} & 71.6 \secondbadcluster{$\downarrow$ 5.9} & 69.3 \secondbadcluster{$\downarrow$ 6.2} & & 53.5 \fourthbadcluster{$\downarrow$ 1.7} & 77.7 \fifthbadcluster{$\downarrow$ 0.3} & 77.3 \fourthcluster{$\uparrow$ 0.6} \\

        GPT 3.5T & 51.3 \secondcluster{$\uparrow$ 2.9} & 82.7 \firstcluster{$\uparrow$ 4.3} & 78.6 \firstcluster{$\uparrow$ 3.1} & & 65.5 \firstcluster{$\uparrow$ 5.8} & 80.9 \firstcluster{$\uparrow$ 3.4} & 77.8 \secondcluster{$\uparrow$ 2.2} & & 57.1 \thirdcluster{$\uparrow$ 1.8} & 79.9 \thirdcluster{$\uparrow$ 2.0} & 79.2 \secondcluster{$\uparrow$ 2.5} \\

        \textsc{TowerInst} 13B & 50.0 \thirdcluster{$\uparrow$ 1.6} & 82.2 \firstcluster{$\uparrow$ 3.8} & 78.7 \firstcluster{$\uparrow$ 3.2} & & 50.7 \secondbadcluster{$\downarrow$ 9.1} & 68.7 \secondbadcluster{$\downarrow$ 8.8} & 66.5 \secondbadcluster{$\downarrow$ 9.0} & & 56.5 \thirdcluster{$\uparrow$ 1.3} & 79.1 \thirdcluster{$\uparrow$ 1.1} & 78.4 \thirdcluster{$\uparrow$ 1.7} \\

        \cdashlinelr{1-12}
        \multicolumn{10}{l}{\textcolor{gray}{\textit{With predicted error spans:}}} 
        \\

        Mixtral 8x7B & 42.9 \secondbadcluster{$\downarrow$ 5.5} & 64.9 \secondbadcluster{$\downarrow$ 13.5} & 58.7 \secondbadcluster{$\downarrow$ 16.8} & & 58.1 \fourthbadcluster{$\downarrow$ 1.6} & 76.4 \fourthbadcluster{$\downarrow$ 1.0} & 73.2 \thirdbadcluster{$\downarrow$ 2.3} & & 51.2 \secondbadcluster{$\downarrow$ 4.1} & 74.4 \secondbadcluster{$\downarrow$ 3.6} & 73.4 \secondbadcluster{$\downarrow$ 3.3} \\
        
        GPT 3.5T & 53.4 \firstcluster{$\uparrow$ 5.0} & 81.6 \firstcluster{$\uparrow$ 3.2} & 77.5 \secondcluster{$\uparrow$ 2.1} & & 63.9 \firstcluster{$\uparrow$ 4.1} & 80.9 \firstcluster{$\uparrow$ 3.5} & 77.9 \secondcluster{$\uparrow$ 2.4} & & 56.2 \fourthcluster{$\uparrow$ 1.0} & 79.1 \thirdcluster{$\uparrow$ 1.1} & 77.9 \thirdcluster{$\uparrow$ 1.1} \\

       \xTower 13B & 52.7 \firstcluster{$\uparrow$ 4.3} & 81.3 \secondcluster{$\uparrow$ 2.9} & 77.0 \thirdcluster{$\uparrow$ 1.5} & & 60.9 \thirdcluster{$\uparrow$ 1.1} & 78.5 \thirdcluster{$\uparrow$ 1.0} & 75.6 \fourthcluster{$\uparrow$ 0.1} & & 56.0 \fourthcluster{$\uparrow$ 0.7} & 79.0 \thirdcluster{$\uparrow$ 1.0} & 78.4 \thirdcluster{$\uparrow$ 1.7} \\

       + Hybrid & 52.4 \firstcluster{$\uparrow$ 4.0} & 82.2 \firstcluster{$\uparrow$ 3.8} & 80.1 \firstcluster{$\uparrow$ 4.6} & & 62.4 \secondcluster{$\uparrow$ 2.6} & 80.0 \secondcluster{$\uparrow$ 2.5} & 78.7 \firstcluster{$\uparrow$ 3.2} & & 55.4 \fourthcluster{$\uparrow$ 0.2} & 79.1 \thirdcluster{$\uparrow$ 1.1} & 78.8 \secondcluster{$\uparrow$ 2.1} \\

        \cdashlinelr{1-12}
        \multicolumn{10}{l}{\textcolor{gray}{\textit{With human-annotated error spans:}}} 
        \\

        Mixtral 8x7B & 42.1 \secondbadcluster{$\downarrow$ 6.2} & 66.8 \secondbadcluster{$\downarrow$ 11.7} & 61.3 \secondbadcluster{$\downarrow$ 14.2} & & 57.7 \thirdbadcluster{$\downarrow$ 2.0} & 76.0 \fourthbadcluster{$\downarrow$ 1.5} & 73.1 \thirdbadcluster{$\downarrow$ 2.4} & & 52.8 \thirdbadcluster{$\downarrow$ 2.5} & 75.7 \thirdbadcluster{$\downarrow$ 2.2} & 74.1 \thirdbadcluster{$\downarrow$ 2.7} \\

        GPT 3.5T & 50.2 \thirdcluster{$\uparrow$ 1.8} & 80.6 \secondcluster{$\uparrow$ 2.2} & 76.5 \fourthcluster{$\uparrow$ 1.0} & & 62.6 \secondcluster{$\uparrow$ 2.8} & 80.0 \secondcluster{$\uparrow$ 2.5} & 77.4 \thirdcluster{$\uparrow$ 1.9} & & 56.5 \thirdcluster{$\uparrow$ 1.3} & 79.2 \thirdcluster{$\uparrow$ 1.2} & 77.9 \thirdcluster{$\uparrow$ 1.2} \\

        \xTower 13B & 50.2 \thirdcluster{$\uparrow$ 1.8} & 81.3 \secondcluster{$\uparrow$ 2.9} & 77.3 \thirdcluster{$\uparrow$ 1.8} & & 60.0 \fourthcluster{$\uparrow$ 0.2} & 77.7 \fourthcluster{$\uparrow$ 0.2} & 75.0 \fifthbadcluster{$\downarrow$ 0.5} & & 56.4 \thirdcluster{$\uparrow$ 1.2} & 79.4 \thirdcluster{$\uparrow$ 1.4} & 78.6 \thirdcluster{$\uparrow$ 1.9} \\

        + Hybrid & 52.7 \firstcluster{$\uparrow$ 4.3} & 82.5 \firstcluster{$\uparrow$ 4.1} & 79.9 \firstcluster{$\uparrow$ 4.4} & & 63.6 \firstcluster{$\uparrow$ 3.8} & 80.8 \firstcluster{$\uparrow$ 3.4} & 79.4 \firstcluster{$\uparrow$ 3.9} & & 56.2 \thirdcluster{$\uparrow$ 1.0} & 79.7 \thirdcluster{$\uparrow$ 1.7} & 79.2 \secondcluster{$\uparrow$ 2.5} \\

        \bottomrule 
    \end{tabular}
    \caption{
    Results for correcting translations with \xComet-predicted or human-annotated error spans. We also show absolute differences from the original translation, where \clustertab{set10-red!15}{red} and \clustertab{set10-blue!15}{blue} denote negative and positive deltas.
    }
    \label{tab:results_correction_referenceless}
\end{table*}

\section{Correcting Translations} 
\label{sec:refining_translations}

Having established that \xTower can provide helpful and useful explanations, we turn to prompting it towards obtaining translation corrections.

\subsection{Experimental Setup}

Our setup for correcting translations follows the same design choices used for explaining translation errors in \S\ref{sec:explaining_translation_errors}, including the test data (WMT23), source of error spans (human vs. \xComet), and prompting format. 
In addition, we employ automatic metrics for measuring translation quality and use robust baselines, which we describe next. We focus on referenceless prompting in the main text.\footnote{In Appendix~\ref{subsec:reference_based_appendix} we carefully investigate the impact of providing reference translations to \xTower.}

\paragraph{Evaluation.} To assess the overall quality of the translation correction produced by LLMs, 
we rely on neural-based metrics, which obtain the strongest results in the WMT23 Metrics shared task~\citep{freitag-etal-2023-results}. 
Specifically, we report \Comet~\citep{rei-etal-2022-comet} as our primary metric, alongside BLEURT~\citep{sellam-etal-2020-bleurt}, and \textsc{CometKiwi}~\citep{rei-etal-2022-cometkiwi}.\footnote{\href{https://huggingface.co/Unbabel/wmt22-comet-da}{\texttt{wmt22-comet-da}} and \href{https://huggingface.co/Unbabel/wmt22-cometkiwi-da}{\texttt{wmt22-cometkiwi-da}}.}
Notably, while the first metrics rely on a reference translation, \textsc{CometKiwi} yields an overall score without a reference. 
We cover lexical metrics in Appendix~\ref{sec:extra_refinement_results}.

\paragraph{Baselines.} We adopt three models as baselines: \TowerInstruct 13B~\citep{alves2024tower}---\Tower's translation-oriented LLM---, Mixtral 8x7B, and GPT 3.5 Turbo.\footnote{We move from GPT 4 to 3.5T due to financial constraints.} 
We use them in a \textbf{translation-only mode}: we pass a source sentence and ask for a translation in a 0-shot manner.
We also prompt Mixtral 8x7B and GPT 3.5T for a translation correction given error spans and explanations.
Since they are not trained to receive this information, we provide a 1-shot example for in-context learning.\footnote{We experiment with 5-shot in Appendix~\ref{sec:extra_refinement_results}, but the results are on par with 1-shot, while also being more costly.}
Appendix~\ref{sec:appendix_prompt_1shot} has more prompt details.

\subsection{Results}

We show results in Table~\ref{tab:results_correction_referenceless} for each language pair.

\paragraph{Is \xTower effective at refining translations?} 
We observe that \xTower's corrections improve the translation quality of the original translations for all language pairs. Interestingly, \xTower obtains similar results with human-annotated and \xComet error spans. For the latter, \Comet deltas vary from 1 to 3 points, \textbf{leading to significant quality improvements for \textsc{en-de} and \textsc{zh-en}.}\footnote{As per \citep{kocmi2024navigating}, \Comet deltas of $\sim$1.0 denote improvements with a 90\% accuracy with human judgments.}

\paragraph{How does \xTower compare to prompting LLMs?} 
Comparing the best scores obtained by \xTower---either from \xComet or human spans---and \TowerInstruct, 
we find that \xTower outperforms \TowerInstruct on \textsc{he-en} and \textsc{zh-en}, with a delta of 9 \Comet points on the former.\footnote{\Tower models were not trained to support Hebrew.}
Interestingly, however, \xTower has a gap of only 0.2 to the original MT for \textsc{he-en}, suggesting that \xTower is only slightly editing the original translation.
Mixtral presents the lowest scores overall, while GPT 3.5T achieves the highest scores overall, outperforming \xTower on all language pairs in terms of BLEURT and \Comet. 
However, in contrast to \xTower, we find that GPT 3.5T displays a consistent drop of performance when refining translations, 
suggesting that it may not utilize error spans and explanations as effectively.

\paragraph{Are the error spans being fixed?}
To assess how effectively the models address the errors highlighted in the prompt, we computed the percentage of fixed error spans with a string matching approach. 
Overall, \xTower fixes 80\% of the errors for \textsc{en-de}, 83\% for \textsc{he-en}, and 84\% for \textsc{zh-en}, while
GPT 3.5T fixes 75\% of the errors for \textsc{en-de}, 82\% for \textsc{he-en}, and 80\% for \textsc{zh-en}. 
These results indicate that both GPT-3.5T and \xTower can, to some degree, leverage error spans and explanations to fix a large portion of the errors, with \xTower showing a consistent edge over GPT-3.5T.

\begin{figure}[t]
    \centering
    \includegraphics[width=\columnwidth]{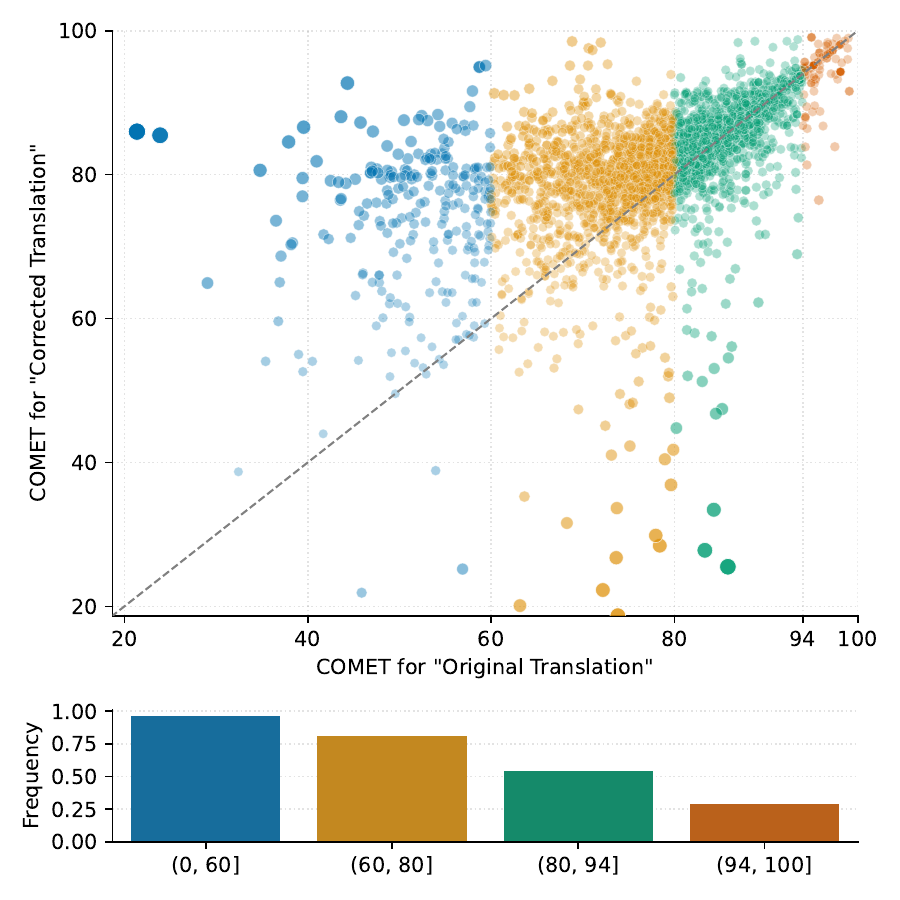}
    \caption{
    At the top, we show the quality of the original translation versus the corrected translation on \textsc{en-de} with human spans. At the bottom, we show how often the latter is higher than the former per quality bin.
    }
    \label{fig:scatter_plot}
\end{figure}

\paragraph{Can we design an effective hybrid approach?}
We have established above that \xTower's corrections can lead to significant improvements in translation quality. Here, we analyse where it is most effective in regards to the translation quality of the original translation. The scatter plot in Figure~\ref{fig:scatter_plot} illustrates the relationship between \Comet scores for original and the corrected translations on \textsc{en-de} samples. It shows that \xTower is most effective for low-quality original translations (\Comet score $\leq$ 80), while for high-quality translations (\Comet score $>$ 80) retaining the original translation may be better.\footnote{This is consistent for all language pairs (\textit{cf.} Figure~\ref{fig:scatter_plots_heen_zh_en}).} 
This is because the test dataset~(WMT23) includes translations from diverse MT systems, including strong models like GPT-4 and (private) commercial systems~\citep{freitag-etal-2023-results}.
Given these findings, we propose a \textbf{hybrid approach} that selects the best method based on the original translation's \Comet score. Instead of a fixed threshold, we find the optimal threshold $\tau$ on 10\% of the samples and use the following rule to obtain the final translation $y$:
\begin{align}
y = \begin{cases} 
y_{\text{original}} & \text{if } m(y_\text{original}) > \tau \\
y_{\text{correction}}  & \text{elif } m(y_\text{correction}) > m(y_\text{original}) \\
y_{\text{original}} & \text{otherwise},
\end{cases} 
\label{eq:hybrid}
\raisetag{47.5pt}
\end{align}
where $m$ is a metric. We use \textsc{CometKiwi}, a \emph{referenceless} metric, as $m$. Results in Table~\ref{tab:results_correction_referenceless} (under "Hybrid") show that this approach consistently improves translation quality across all language pairs, with boosts as high as 2 \Comet points for \textsc{he-en}.
These results suggest that a hybrid approach can significantly improve translation performance, especially for the more realistic scenario of using \xComet spans, while also reducing inference costs by only querying \xTower sporadically.\footnote{
Portion of original translations kept: \{46\%, 41\%\}  for \textsc{en-de}, \{49\%, 48\%\} for \textsc{he-en}, and \{30\%, 32\%\} for \textsc{zh-en} using \xComet and human spans, respectively.}

\begin{figure}[t]
    \centering
    \includegraphics[width=\columnwidth]{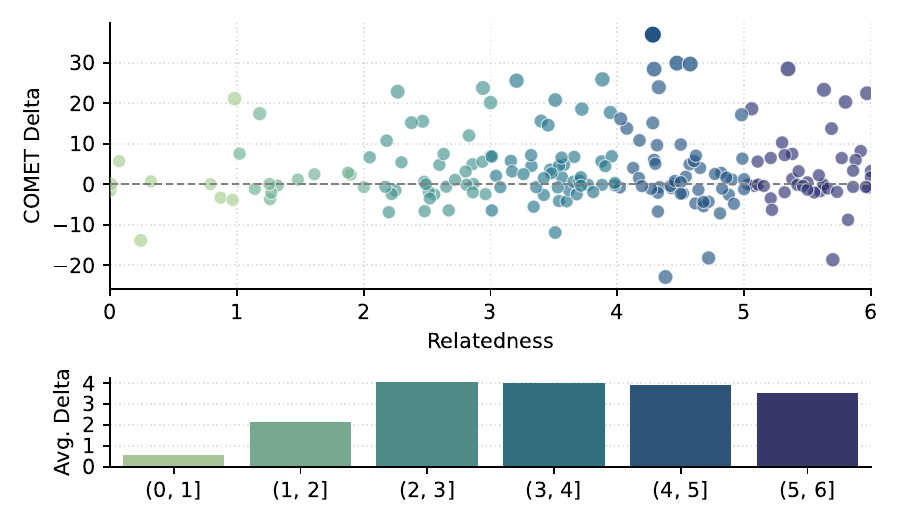}
    \vspace*{-0.25cm}
    \caption{Delta between \Comet scores for corrected and  original translations according to how related explanations are to error spans.}
    \label{fig:delta_relatedness_plot}
\end{figure}

\paragraph{How does explanation quality affect corrections?}
In Figure~\ref{fig:delta_relatedness_plot} we show that the largest quality gains are typically associated with explanations that have a high relatedness score (\S\ref{sec:relatedness}).
Furthermore, we find a negative Pearson correlation~($r = -0.15$) between explanations' relatedness and original translations' \Comet scores, 
highlighting that higher quality explanations are often associated with poorer quality original translations.
This suggests that \textbf{high-quality explanations lead to significant improvements primarily for contexts where the initial translation quality is poor}, as hinted by Figure~\ref{fig:scatter_plot}.

\section{Related Work}

Here, we discuss key related works in the domains of free-text explanations, automatic post-editing, span-level error detection, and the use of LLMs for translation and error explanation.

\paragraph{Free-text Explanations.}
Recent work has explored methods for generating free-text explanations either by utilizing human-written examples~\citep{marasovic-etal-2022-shot,wiegreffe-etal-2022-reframing} or by prompting LLMs~\citep{wei2022chain,jung-etal-2022-maieutic,atanasova-etal-2023-faithfulness,joshi-etal-2023-machine}. 
However, these explanations are typically produced to understand a model's decision rather than being constrained to justify marked spans in the input. In a similar vein, \citet{feldhus-etal-2023-saliency} propose leveraging dense saliency maps to improve the verbalization of explanations by LLMs. 
In contrast, \xTower focuses on producing explanations that are tied to specific error spans (\S\ref{sec:relatedness}) and helpful to humans (\S\ref{sec:helpfulness}), within the context of MT.

\paragraph{Span-level Error Detection and Correction.}
In the context of span-level error detection, \textsc{AutoMQM}~\citep{fernandes-etal-2023-devil}, InstructScore~\citep{xu-etal-2023-instructscore}, and \xComet~\citep{guerreiro2023xcomet} have demonstrated the effectiveness of using neural models to identify errors in machine translations. 
For correcting errors in translations, a task more generally known as automatic post-editing (APE; \citealt{simard2007statistical,bhattacharyya-etal-2023-findings}), recent works prompt LLMs to produce suggestions for a new translation, such as \textsc{TowerAPE}~\citep{alves2024tower} and prompting GPT-4~\citep{raunak-etal-2023-leveraging}.
We experiment with error spans annotated by humans or predicted by \xComet for correcting translations in \S\ref{sec:refining_translations}. 
Finally, incorporating detailed error feedback into post-editing prompts has been concurrently explored by \citet{ki2024guiding,xu-etal-2024-llmrefine}. 
In particular, LLMRefine~\citep{xu-etal-2024-llmrefine} casts fine-grained error feedback as ``succinct explanations,'' guiding the model towards improved translations through iterative refinements. 
While effective, their approach sidestep comprehensive explanations, which, as shown in \S\ref{sec:refining_translations}, can further improve the translation correction process.

\paragraph{LLMs for Translating and Explaining Translation Errors.} LLMs have been increasingly employed for translation tasks. \Tower~\citep{alves2024tower} and ALMA~\citep{xu2024alma} are notable examples of models designed specifically for translation-related tasks. 
InstructScore, a recent work by \citet{xu-etal-2023-instructscore}, uses LLMs to provide explanations for translation errors. 
However, in contrast to \xTower, InstructScore relies on reference translations, sidesteps the information in the source sentence, and produces explanations only as a by-product to improve quality score predictions. 
Additionally, while InstructScore focuses on producing a single quality score to reflect overall translation quality, \xTower not only provides plausible and helpful explanations for humans, but also generates translation corrections.

\section{Conclusions}

In this paper, we introduced \xTower, a multilingual LLM designed to provide free-text explanations for translation errors and generate corrected translations. 
By leveraging the strengths of \Tower and integrating specialized error detection from \xComet, 
\xTower can improve the interpretability of machine translation outputs in an automatic process. 
Our evaluations demonstrate that \xTower not only produces high-quality and helpful explanations, as assessed by human evaluation, but can also significantly improves translation quality, especially when combined with accurate error spans.
Furthermore, we propose a hybrid approach that dynamically selects between using the original translation or querying \xTower for a correction, resulting in overall improvements in translation quality for all language pairs.

\section*{Acknowledgments}

We thank Patrick Fernandes, António Farinhas, and Duarte Alves, Dennis Fucci, and Taisiya Glushkova for their useful and constructive comments. 
We also thank the human annotators for their help in evaluating explanations.
This work was supported by the Portuguese Recovery and Resilience Plan through project C645008882-00000055 (Center for ResponsibleAI), by the EU’s Horizon Europe Research and Innovation Actions (UTTER, contract 101070631), by the project DECOLLAGE (ERC-2022-CoG 101088763), and by Fundação para a Ciência e Tecnologia through contract UIDB/50008/2020.

\section*{Limitations}

While \xTower significantly advances machine translation interpretability, it has various limitations. Even though the model's dependence on external error span detector tools like \xComet brings modularity and flexibility, it also introduces pipeline complexity. Our evaluation, focused on the few language pairs which have MQM annotations available, may not generalize across all languages and domains. Additionally, the computational resources required for distillation and finetuning are substantial, limiting reproducibility for some users. The generated explanations, though helpful, may not always faithfully represent the model's reasoning or effectively guide users. Lastly, potential biases in the training data could affect translation and explanation quality, requiring further work to ensure fairness and reliability. 

\section*{Potential Risks}

The use of \xTower may carry potential risks. One concern is the possibility of the model generating fluent but misleading explanations, which could affect user trust. 
There are also fairness considerations; as discussed above, the model might inadvertently reinforce biases present in the training data, potentially disadvantaging historically marginalized groups. 
Lastly, the focus on certain languages and datasets could lead to the underrepresentation of less commonly spoken languages.
Careful monitoring and ongoing evaluation, such as detecting and overcoming hallucinations~\citep{guerreiro-etal-2023-looking,dale-etal-2023-detecting}, can help mitigate these risks and ensure the model's responsible use.

\bibliography{anthology_small,custom}

\newpage

\appendix

\section{Datasets Statistics}
\label{sec:datasets_statistics}

We show statistics for all datasets used in this work in Table~\ref{tab:data_stats}.

\begin{table}[t]
    \centering
    \small
    \setlength{\tabcolsep}{4pt}
    \begin{tabular}{lllll}
    \toprule
                    & \textsc{en-de} & \textsc{he-en} & \textsc{zh-en} & \textsc{en-ru} \\
    \midrule
    \textit{WMT 2022} \\
    \# Samples          & 8,815 & - & 13,631 & 10,996 \\
    \# Error Spans      & 14,174 & - & 26,506 & 22,508 \\
    Avg. Input Length   & 42.2 & - & 72.3 & 40.5 \\
    Avg. Span Length    & 1.8 & - & 2.4 & 1.6 \\
    \cdashlinelr{1-5}
    \textit{WMT 2023} \\
    \# Samples          & 4,111 & 5,325 & 15,690 & - \\
    \# Error Spans      & 17,439 & 8,476 & 43,649 & - \\
    Avg. Input Length   & 190.0 & 18.1 & 52.9 & - \\
    Avg. Span Length    & 3.0 & 1.0 & 2.5 & - \\
    \cdashlinelr{1-5}
    \multicolumn{5}{l}{\textcolor{gray}{\textit{\xComet spans (without references):}}} 
    \\
    \# Error Spans      & 33,774 & 16,816 & 57,917 & - \\
    Avg. Span Length    & 2.4 & 1.1 & 2.0 & - \\
    
    \cdashlinelr{1-5}
    \multicolumn{5}{l}{\textcolor{gray}{\textit{\xComet spans (with references):}}} 
    \\
    \# Error Spans      & 30,856 & 16,434 & 53,602 & - \\
    Avg. Span Length    & 2.5 & 1.1 & 2.0 & - \\
    
    \bottomrule
    \end{tabular}
    \caption{Datasets statistics.}
    \label{tab:data_stats}
\end{table}

\section{Prompting}
\label{sec:appendix_prompt_1shot}

\paragraph{Prompting explanations and translation correction.} For 1-shot, we pass a unique example as input: for \textsc{en-de} we pass a single \textsc{en-de} example, whereas for \textsc{he-en} and \textsc{zh-en} we pass a \textsc{zh-en} example. For 5-shot, we pass a list of 5 examples containing 3 \textsc{en-de}, 1 \textsc{en-ru}, and 1 \textsc{zh-en} samples. 
For all models, we sample new tokens using a temperature set to zero.
We provide an example of our prompt template used for 1-shot \textsc{en-de} experiments in Table~\ref{tab:example_prompt_gpt4}.

\paragraph{Prompting translation-only LLMs.} For the translation LLMs baselines, we use the prompt shown in Table~\ref{tab:prompt_transaltion_0shot} to obtain translations.

\begin{table}[!htb]
    \small
    \centering
    \begin{tabular}{m{0.95\columnwidth}}
    \toprule
    Translate the following English source text to German: \\
    \cdashlinelr{1-1}
    \textbf{English source:} This is a great product and suitable for all bikes, cars and commercial applications. \\
    \cdashlinelr{1-1}
    \textbf{German translation:} Dieses großartige Produkt eignet sich für alle Motorräder, Autos und gewerbliche Anwendungen. \\
    \bottomrule
    \end{tabular}
    \caption{0-shot prompt for generating translations.}
    \label{tab:prompt_transaltion_0shot}
\end{table}

\section{Human Evaluation}
\label{sec:human_annotation}

\begin{figure*}[!htb]
    \centering
    \includegraphics[width=\textwidth]{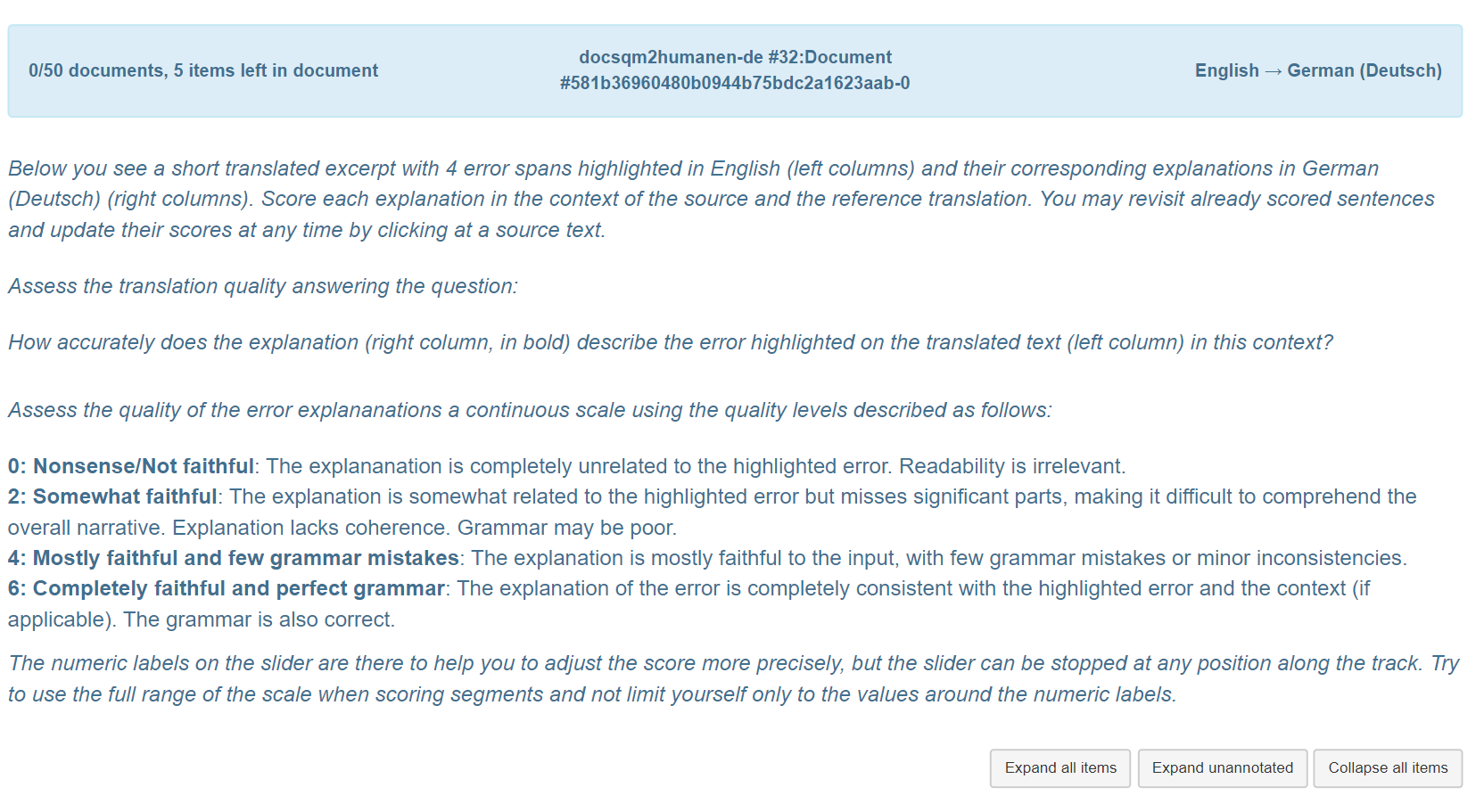}%
    \vspace{0.2cm}
      \includegraphics[width=\textwidth]{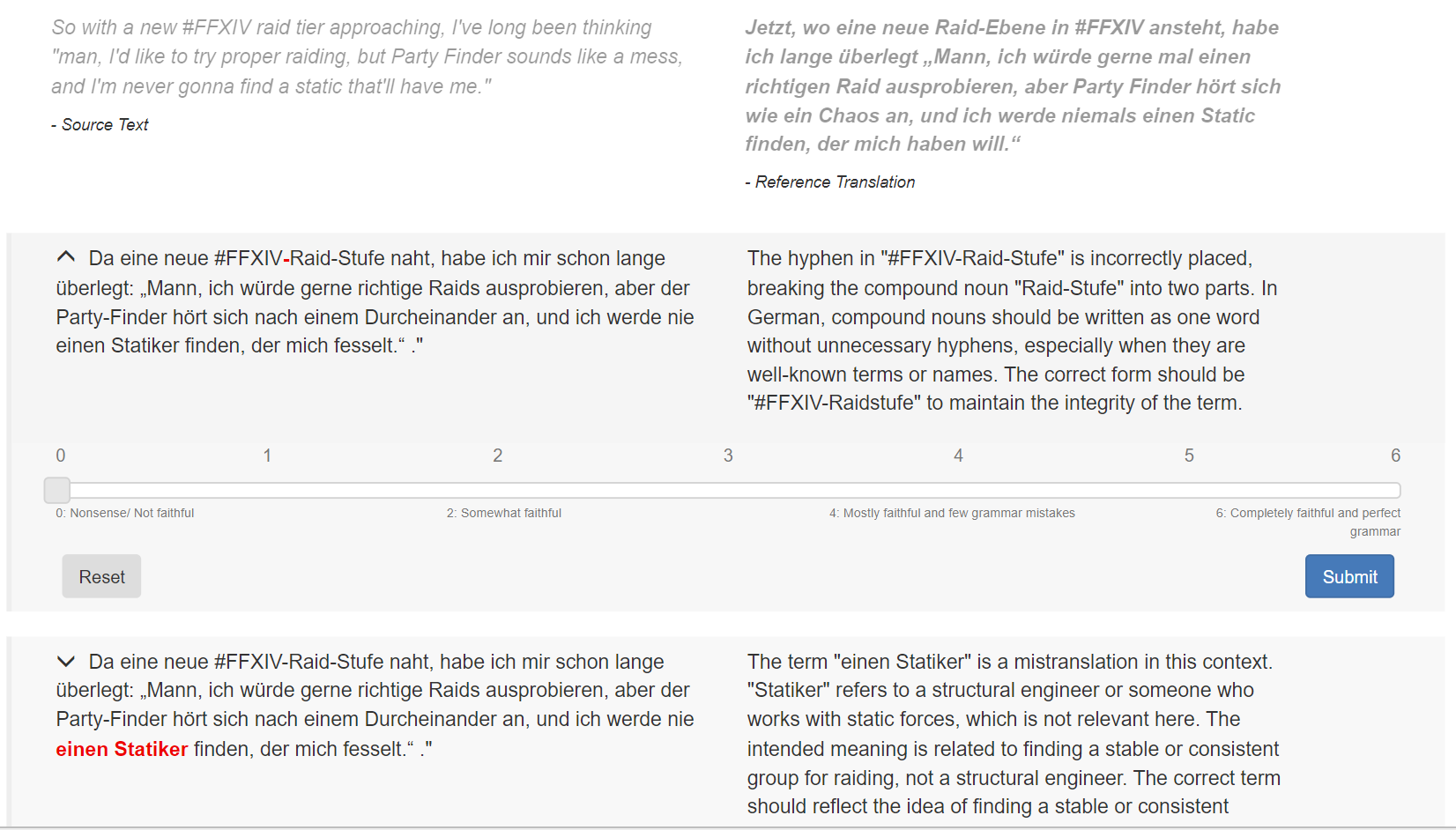}%
    \caption{Screenshot of the \textit{relatedness} task interface presented to annotators. }
    \label{fig:task_screenshot_relate}
\end{figure*}

\begin{figure*}[!htb]
    \centering
      \includegraphics[width=\textwidth]{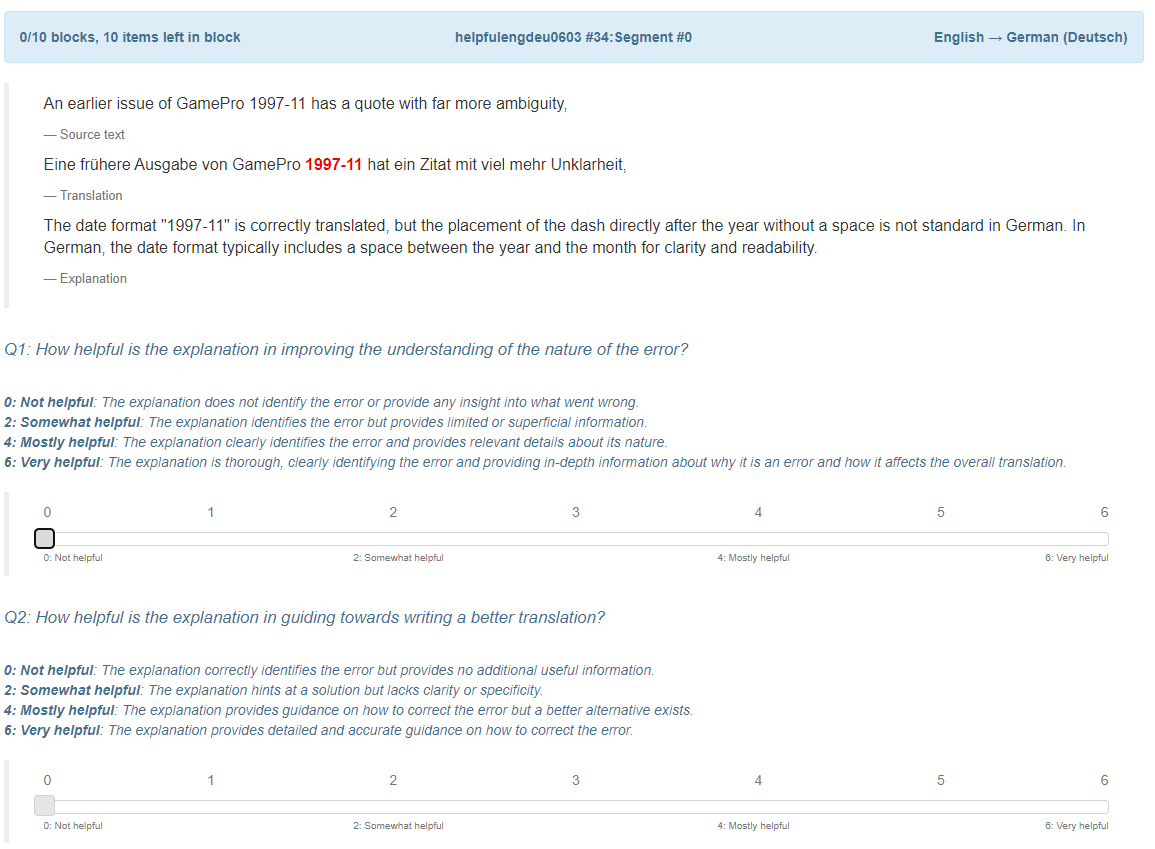}%
    \caption{Screenshot of the \textit{helpfulness} task interface presented to annotators. }
    \label{fig:task_screenshot_help}
\end{figure*}

\paragraph{Detailed Task Instructions.} We present the detailed task instructions provided to the annotators in Figures~\ref{fig:task_screenshot_relate} (\textit{relatedness}) and ~\ref{fig:task_screenshot_help} (\textit{helpfulness}). The interface was created using Appraise \cite{federmann-2018-appraise}.

\paragraph{Inter-annotator agreement.} We ask human annotators to assess the translations at both the explanation and document levels. The inter-annotator agreement was measured using two statistical metrics: Pearson correlation coefficient ($r$) and Spearman rank correlation coefficient ($\rho$). 
Specifically, following \citet{pavlick-tetreault-2016-empirical}, for each instance either at explanation or document-level, we randomly choose one annotator's scores to be the scores provided by Annotator 1, and take the mean scores of the other two annotators to be the scores given by an Annotator 2. 
We then compute the correlation for these two simulated annotators. 
Table \ref{tab:annotators_agreement} presents the results. 
The results indicate that while human annotators exhibit higher consistency for \textsc{en-de} translations, the agreement is generally lower for \textsc{zh-en} translations. For \xComet spans, however, annotators agree more consistently across both language pairs.

\begin{table}[t]
    \small
    \centering
    \setlength{\tabcolsep}{4pt}
    \begin{tabular}{ll cc c cc}
    \toprule
    & & \multicolumn{2}{c}{\sc en-de} & & \multicolumn{2}{c}{\sc zh-en} \\
    \cmidrule(lr){3-4} \cmidrule(lr){6-7} 
    \sc Span & \sc Level & $r$ & $\rho$ & & $r$ & $\rho$ \\
    \midrule
    
    Human & Explanation  & 0.56 & 0.51 & & 0.34 & 0.20 \\
    Human & Document     & 0.50 & 0.38 & & 0.21 & 0.17 \\
    
    \xComet & Explanation & 0.47 & 0.46 & & 0.54 & 0.46 \\
    \xComet & Document    & 0.39 & 0.40 & & 0.50 & 0.46 \\

    \cdashlinelr{1-7}
    \it Average & Explanation & 0.52 & 0.48 & & 0.44 & 0.33 \\
    \it Average & Document & 0.45 & 0.39 & & 0.35 & 0.32 \\
    \bottomrule
    \end{tabular}
    \caption{Inter-annotator agreement at explanation and document-level, according to Pearson's $r$ and Spearman's $\rho$ correlation coefficients.}
    \label{tab:annotators_agreement}
\end{table}

\paragraph{Sample size.} For relatedness experiments, we evaluated a total of 282 explanations for \textsc{en-de} and
279 for \textsc{zh-en} (561 in total). For helpfulness, we evaluated 83 explanations \textsc{en-de} and
99 for \textsc{zh-en} (182 in total).

\paragraph{Participants Details.} We hired native speakers of Chinese and German (fluent in English) for this task (four females and two males). They were compensated at \$24 per hour.

\section{Translation Correction}
\label{sec:extra_refinement_results}

\subsection{Referenceless}

\paragraph{K-shot prompt.} 
In Table~\ref{tab:k_shot_results}, we present results with $k$-shot for samples with human-annotated translation error spans in terms of \Comet, covering both referenceless and reference-based setups. 
Note that only xTower was evaluated with $k=0$, as it was finetuned on explanations, and thus it can sidestep the in-context learning examples. 
Mixtral 8x7B seems to benefit more with $k=5$ than other models for \textsc{he-en} and \textsc{zh-en}, but looses more around 4 \Comet points for \textsc{en-de}.
On the other hand, GPT 3.5T performs better with $k=1$ than with $k=5$ for referenceless experiments, with $k=1$ results also being very close to $k=5$ for reference-based experiments.
Finally, xTower with $k=5$ usually obtains slightly better results than with $k \in \{0, 1\}$ (delta within 0.2-0.4), but it introduces substantial runtime and memory costs as the prompt grows $\sim$5 times its original size. These findings motivated us to select $k=1$ for Mixtral and GPT, and $k=0$ for xTower, for all experiments in the paper.

\begin{table}[t]
    \small
    \centering
    \setlength{\tabcolsep}{3.3pt}
    \begin{tabular}{ll ccc c@{\ } ccc}
    \toprule
    & & \multicolumn{3}{c}{Referenceless} & & \multicolumn{3}{c}{Reference-based} \\
    \cmidrule(lr){3-5} \cmidrule(lr){6-9} 
    \sc Model & $k$ & 
    en-de & he-en & zh-en & &
    en-de & he-en & zh-en \\
    \midrule
    Mixtral    & 1     & 66.8 & 76.0 & 75.7 & & 69.9 & 84.5 & 80.0 \\
    Mixtral    & 5     & 63.1 & 77.1 & 76.6 & & 65.4 & 85.3 & 81.4 \\
    GPT 3.5T    & 1     & 80.6 & 80.0 & 79.2 & & 83.7 & 87.6 & 82.9 \\
    GPT 3.5T    & 5     & 79.9 & 79.9 & 79.0 & & 81.4 & 87.9 & 83.2 \\
    xTower     & 0     & 81.3 & 77.7 & 79.4 & & 84.1 & 88.2 & 83.6 \\
    xTower     & 1     & 81.4 & 77.6 & 79.3 & & 84.3 & 88.7 & 83.9 \\
    xTower     & 5     & 81.2 & 77.9 & 79.2 & & 84.4 & 88.6 & 84.0 \\
    \bottomrule
    \end{tabular}
    \caption{Results for translation refinement with $k$-shot prompting in terms of \Comet.}
    \label{tab:k_shot_results}
\end{table}

\subsection{Reference-based}
\label{subsec:reference_based_appendix}

For many use cases, users can provide an initial translation draft and then query \xTower with the goal of obtaining an improved version. Here, we investigate the impact of providing a reference translation to the input on the quality of corrected translations.

\subsubsection{Experimental Setup}

\paragraph{Distillation data.} 
Since references might play an important role in understanding and explaining translation errors, for example by offering context and highlighting specific areas where the translation deviates from the ideal, we include the reference in our prompts in 50\% of the cases during distillation. 
Consequently, after finetuning, this approach allows us to balance between leveraging references for better explanations and ensuring the model engages in genuine error correction.

\paragraph{Prompting with \xComet spans.} 
Since we introduce the reference translation as an additional signal to our prompt,
we rerun \xComet with source-translation-reference triplets as input, obtaining a total of 99,892 spans. 

\paragraph{Hybrid strategy.} We use the same hybrid approach use for reference-less experiments, as defined in Equation~\ref{eq:hybrid}. However, here we use \Comet as $m$, a reference-based metric.

\subsubsection{Results}

We present our results in Table~\ref{tab:results_correction_referencebased}. Next, we discuss out main findings.

\paragraph{What's the gap to referenceless?}
Comparing the results with and without references, we find that reference-based models consistently outperform referenceless ones across all metrics and language pairs. 
For example, we obtain \Comet boosts of roughly 3 points for \textsc{en-de}, 11 for \textsc{he-en}, and 5 for \textsc{zh-en}.
Moreover, we note that human-annotated spans yield again similar results with \xComet spans across the board. 
These findings indicate that \xTower effectively leverages references, leading to significant improvements for the task of correcting translations.

\begin{table}[t]
    \small
    \centering
    \setlength{\tabcolsep}{3.5pt}
    \begin{tabular}{l ccc c@{\ } ccc c@{\ } ccc}
        \toprule
        & \multicolumn{3}{c}{\sc en-de} & & 
        \multicolumn{3}{c}{\sc he-en} & & 
        \multicolumn{3}{c}{\sc zh-en} \\ 
        \cmidrule{2-4} \cmidrule{6-8} \cmidrule{10-12} 
        \sc Spans & 
        $C$ & $S$ & $\Delta$ & &
        $C$ & $S$ & $\Delta$ & &
        $C$ & $S$ & $\Delta$ \\
        \midrule
        \multicolumn{10}{l}{\textcolor{gray}{\textit{Without references:}}} \\
        \sc xComet   & .01 & .42 & .44 & & .01 & .41 & .39 & & .01 & .17 & .77 \\
        \sc Human    & .01 & .43 & .49 & & .00 & .40 & .39 & & .01 & .18 & .77 \\
        \cdashlinelr{1-12}
        \multicolumn{10}{l}{\textcolor{gray}{\textit{With references:}}} \\
        \sc xComet   & .10 & .57 & .49 & & .18 & .72 & .51 & & .08 & .33 & .79 \\
        \sc Human    & .06 & .57 & .55 & & .15 & .69 & .53 & & .07 & .33 & .78 \\
        \bottomrule
    \end{tabular}
    \caption{Portion of samples where the corrected translation is same as the reference ($C\downarrow$), their normalized Levenshtein similarity ($S\downarrow$), and how often the former is judged better than the latter by \textsc{CometKiwi} ($\Delta\uparrow$).}
    \label{tab:reference_copy}
\end{table}

\paragraph{Is \xTower simply copying the reference?} Since we are now providing a reference translation to \xTower, it is not clear whether the quality gap that we have measured is not just an effect of copying the provided reference. To address this question, we computed two additional metrics: the percentage of translation corrections that are identical to the reference, and their closeness using normalized Levenshtein similarity. The results, presented in Table~\ref{tab:reference_copy}, indicate that \xTower does not simply copy the reference. While the translation corrections become more similar to the reference, this is beneficial as it shows the model relies on the reference to generate improved translations.
Furthermore, to determine the quality of these improvements, we compared the \textsc{CometKiwi} scores of the corrected translations and the original references relative to the source.
The results show that this percentage is generally above 50\%, demonstrating that \xTower effectively produces translations that are on par with or better than the original references.

\paragraph{Is the hybrid approach effective?} 
Our hybrid approach, which dynamically alternates between utilizing high-quality original translations and high-quality corrections, yields significant improvements, particularly in terms of \Comet and BLEURT scores, just as observed in referenceless experiments in \S\ref{sec:refining_translations}. 
Overall, these findings highlight the full potential of \xTower towards improving translation quality.

\begin{table*}[t]
    \small
    \centering
    \setlength{\tabcolsep}{3pt}
    \begin{tabular}{l lll c@{\ \ } lll c@{\ \ } lll}
        \toprule
        & \multicolumn{3}{c}{\sc en-de} & & 
        \multicolumn{3}{c}{\sc he-en} & & 
        \multicolumn{3}{c}{\sc zh-en} \\ 
        \cmidrule{2-4} \cmidrule{6-8} \cmidrule{10-12} 
        \sc Model & 
        \sc bleurt & \sc comet & \sc ckiwi & &
        \sc bleurt & \sc comet & \sc ckiwi & &
        \sc bleurt & \sc comet & \sc ckiwi \\
        \midrule

        Original MT & 48.4 \othercluster{0.0} & 78.4 \othercluster{0.0} & 75.5 \othercluster{0.0} & & 59.8 \othercluster{0.0} & 77.5 \othercluster{0.0} & 75.5 \othercluster{0.0} & & 55.2 \othercluster{0.0} & 78.0 \othercluster{0.0} & 76.7 \othercluster{0.0} \\

        \cdashlinelr{1-12}
        \multicolumn{10}{l}{\textcolor{gray}{\textit{Translation-only LLMs:}}} \\

        Mixtral 8x7B & 46.4 \fourthbadcluster{$\downarrow$ 2.0} & 80.4 \thirdcluster{$\uparrow$ 2.0} & 76.6 \thirdcluster{$\uparrow$ 1.1} & & 53.9 \secondbadcluster{$\downarrow$ 5.9} & 71.6 \secondbadcluster{$\downarrow$ 5.9} & 69.3 \secondbadcluster{$\downarrow$ 6.2} & & 53.5 \fourthbadcluster{$\downarrow$ 1.7} & 77.7 \fifthbadcluster{$\downarrow$ 0.3} & 77.3 \fourthcluster{$\uparrow$ 0.6} \\

        GPT 3.5T & 51.3 \secondcluster{$\uparrow$ 2.9} & 82.7 \firstcluster{$\uparrow$ 4.3} & 78.6 \firstcluster{$\uparrow$ 3.1} & & 65.5 \firstcluster{$\uparrow$ 5.8} & 80.9 \firstcluster{$\uparrow$ 3.4} & 77.8 \secondcluster{$\uparrow$ 2.2} & & 57.1 \thirdcluster{$\uparrow$ 1.8} & 79.9 \thirdcluster{$\uparrow$ 2.0} & 79.2 \secondcluster{$\uparrow$ 2.5} \\

        \textsc{TowerInst} 13B & 50.0 \thirdcluster{$\uparrow$ 1.6} & 82.2 \firstcluster{$\uparrow$ 3.8} & 78.7 \firstcluster{$\uparrow$ 3.2} & & 50.7 \secondbadcluster{$\downarrow$ 9.1} & 68.7 \secondbadcluster{$\downarrow$ 8.8} & 66.5 \secondbadcluster{$\downarrow$ 9.0} & & 56.5 \thirdcluster{$\uparrow$ 1.3} & 79.1 \thirdcluster{$\uparrow$ 1.1} & 78.4 \thirdcluster{$\uparrow$ 1.7} \\

        \cdashlinelr{1-12}
        \multicolumn{10}{l}{\textcolor{gray}{\textit{With predicted error spans:}}} 
        \\

        Mixtral 8x7B & 49.4 \thirdcluster{$\uparrow$ 1.0} & 70.0 \secondbadcluster{$\downarrow$ 8.4} & 62.8 \secondbadcluster{$\downarrow$ 12.7} & & 74.1 \firstcluster{$\uparrow$ 14.4} & 85.6 \firstcluster{$\uparrow$ 8.2} & 77.2 \thirdcluster{$\uparrow$ 1.7} & & 62.4 \firstcluster{$\uparrow$ 7.1} & 80.6 \secondcluster{$\uparrow$ 2.6} & 75.2 \fourthbadcluster{$\downarrow$ 1.5} \\

        GPT 3.5T & 63.3 \firstcluster{$\uparrow$ 15.0} & 85.2 \firstcluster{$\uparrow$ 6.8} & 78.6 \firstcluster{$\uparrow$ 3.1} & & 80.2 \firstcluster{$\uparrow$ 20.5} & 88.8 \firstcluster{$\uparrow$ 11.4} & 79.3 \firstcluster{$\uparrow$ 3.8} & & 66.5 \firstcluster{$\uparrow$ 11.2} & 83.3 \firstcluster{$\uparrow$ 5.3} & 77.7 \fourthcluster{$\uparrow$ 0.9} \\

        xTower 13B & 62.9 \firstcluster{$\uparrow$ 14.6} & 84.6 \firstcluster{$\uparrow$ 6.2} & 77.7 \secondcluster{$\uparrow$ 2.2} & & 80.5 \firstcluster{$\uparrow$ 20.8} & 89.0 \firstcluster{$\uparrow$ 11.5} & 79.3 \firstcluster{$\uparrow$ 3.8} & & 66.8 \firstcluster{$\uparrow$ 11.6} & 83.7 \firstcluster{$\uparrow$ 5.7} & 78.2 \thirdcluster{$\uparrow$ 1.5} \\

        + Hybrid & 62.4 \firstcluster{$\uparrow$ 14.0} & 85.8 \firstcluster{$\uparrow$ 7.4} & 79.4 \firstcluster{$\uparrow$ 3.9} & & 80.2 \firstcluster{$\uparrow$ 20.4} & 88.4 \firstcluster{$\uparrow$ 10.9} & 79.5 \firstcluster{$\uparrow$ 4.0} & & 66.5 \firstcluster{$\uparrow$ 11.2} & 83.8 \firstcluster{$\uparrow$ 5.8} & 78.0 \thirdcluster{$\uparrow$ 1.3} \\

        \cdashlinelr{1-12}
        \multicolumn{10}{l}{\textcolor{gray}{\textit{With human-annotated error spans:}}} 
        \\

        Mixtral 8x7B & 46.3 \thirdbadcluster{$\downarrow$ 2.1} & 69.9 \secondbadcluster{$\downarrow$ 8.5} & 63.5 \secondbadcluster{$\downarrow$ 12.0} & & 72.1 \firstcluster{$\uparrow$ 12.3} & 84.5 \firstcluster{$\uparrow$ 7.1} & 76.5 \fourthcluster{$\uparrow$ 0.9} & & 60.7 \firstcluster{$\uparrow$ 5.4} & 80.0 \secondcluster{$\uparrow$ 2.0} & 75.3 \fourthbadcluster{$\downarrow$ 1.4} \\
        
        GPT 3.5T & 58.5 \firstcluster{$\uparrow$ 10.2} & 83.7 \firstcluster{$\uparrow$ 5.3} & 77.9 \secondcluster{$\uparrow$ 2.5} & & 77.7 \firstcluster{$\uparrow$ 18.0} & 87.6 \firstcluster{$\uparrow$ 10.2} & 78.7 \firstcluster{$\uparrow$ 3.2} & & 65.3 \firstcluster{$\uparrow$ 10.0} & 82.9 \firstcluster{$\uparrow$ 4.9} & 77.9 \thirdcluster{$\uparrow$ 1.2} \\

        xTower 13B & 59.1 \firstcluster{$\uparrow$ 10.7} & 84.1 \firstcluster{$\uparrow$ 5.7} & 77.8 \secondcluster{$\uparrow$ 2.4} & & 78.8 \firstcluster{$\uparrow$ 19.1} & 88.2 \firstcluster{$\uparrow$ 10.8} & 78.7 \firstcluster{$\uparrow$ 3.2} & & 66.5 \firstcluster{$\uparrow$ 11.2} & 83.6 \firstcluster{$\uparrow$ 5.6} & 78.5 \thirdcluster{$\uparrow$ 1.8} \\

        + Hybrid & 61.7 \firstcluster{$\uparrow$ 13.3} & 86.0 \firstcluster{$\uparrow$ 7.6} & 79.7 \firstcluster{$\uparrow$ 4.2} & & 79.6 \firstcluster{$\uparrow$ 19.8} & 88.6 \firstcluster{$\uparrow$ 11.1} & 80.1 \firstcluster{$\uparrow$ 4.5} & & 67.1 \firstcluster{$\uparrow$ 11.9} & 84.3 \firstcluster{$\uparrow$ 6.3} & 78.5 \thirdcluster{$\uparrow$ 1.8} \\
        
        \bottomrule 
    \end{tabular}
    \caption{Reference-based results for correcting translations conditioned on explanations and error spans predicted via \xComet or obtained via human annotation. We also show the absolute difference to the original translation, with \clustertab{set10-red!15}{red} and \clustertab{set10-blue!15}{blue} denoting negative and positive deltas, respectively. 
    }
    \label{tab:results_correction_referencebased}
\end{table*}

\subsection{Additional Results}
\label{sec:additional_refinement_results}

\paragraph{Lexical metrics.} For completeness, we include lexical metrics for referenceless and reference-based experiments for the translation refinement task in Table~\ref{tab:full_results_refinement_appendix}. Specifically, we include BLEU and ChrF.\footnote{SacreBLEU signature: \texttt{|1|mixed|no|13a|exp|}.}

\paragraph{\TowerBase vs \xTower.}  
To verify whether \xTower maintains the original \TowerBase translation capabilities after extending it, we also report its performance as a translation-only LLM in Table~\ref{tab:full_results_refinement_appendix}. 
That is, we prompt \xTower with the 0-shot template shown in Table~\ref{tab:prompt_transaltion_0shot}. 
The table shows that \xTower performs on par or slightly surpass \TowerInstruct for all language pairs in terms of BLEURT, \Comet, and \textsc{CometKiwi}. 
This suggests that \xTower not only keeps the original translation capabilities of \TowerBase, but also holds potential to improve them.

\begin{figure*}[!htb]
    \centering
    \includegraphics[width=0.49\textwidth]{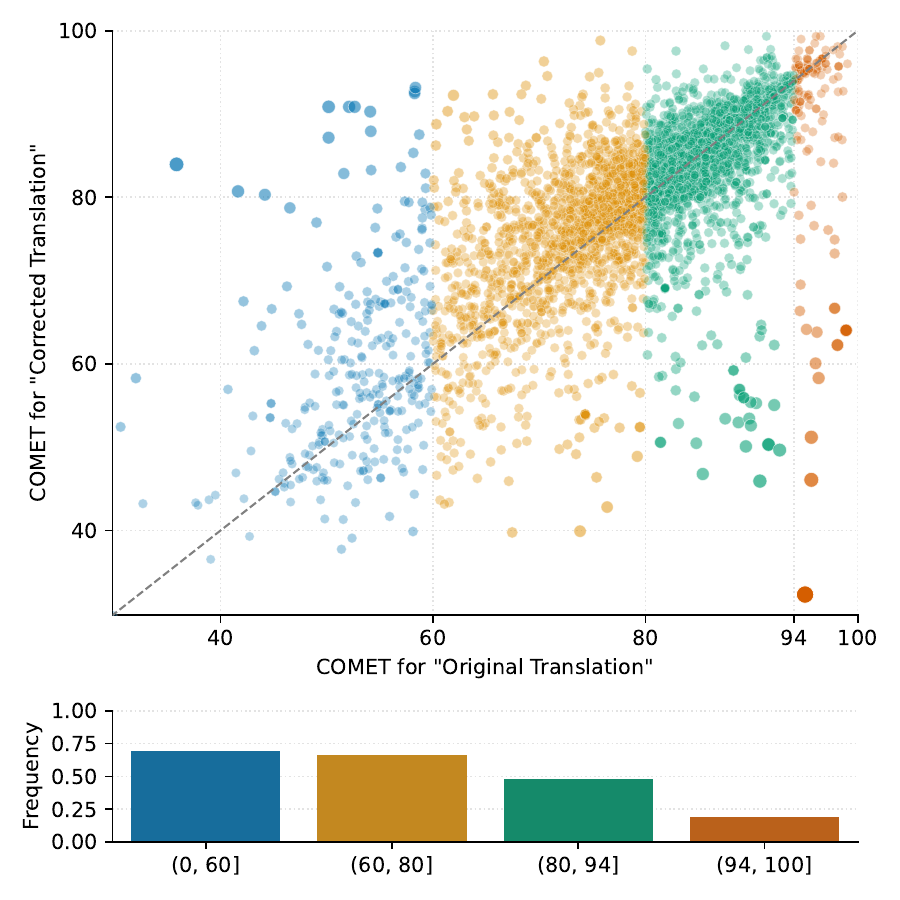}%
    \hfill
    \includegraphics[width=0.49\textwidth]{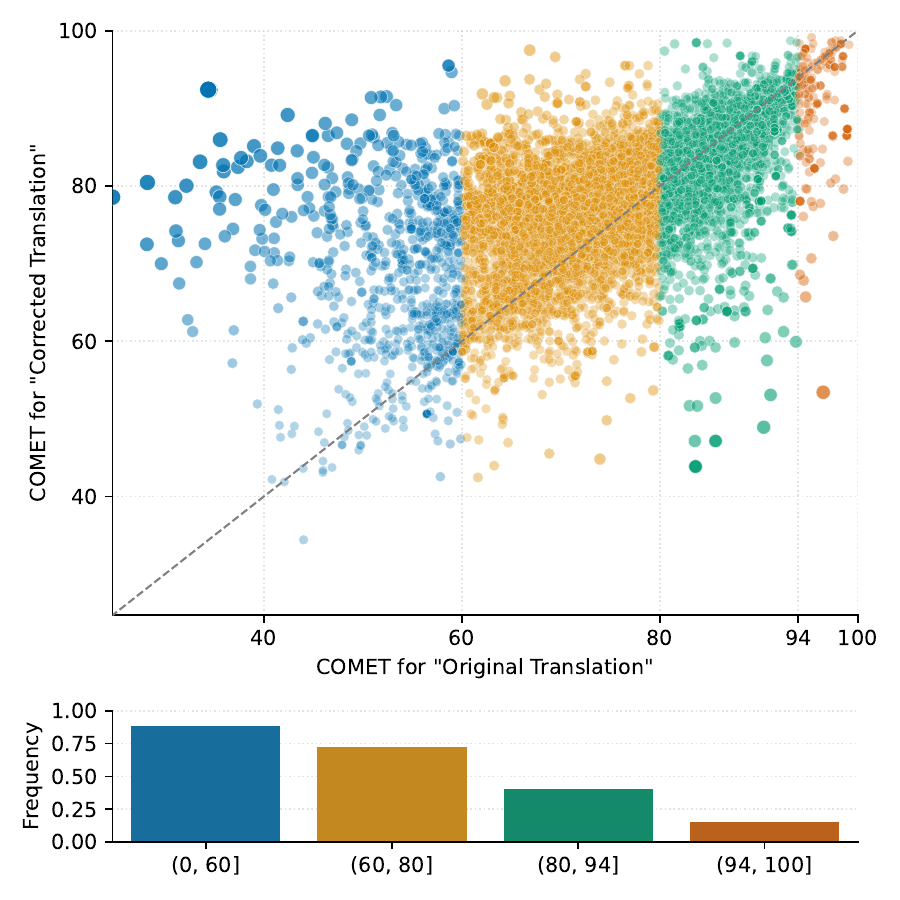}
    \caption{\Comet of the original translation versus the corrected translation with human-annotated spans for \textsc{he-en} (left) and \textsc{zh-en} (right). 
    At the bottom, we show how often the \Comet for the corrected translation is higher than for the original per quality bin.}
    \label{fig:scatter_plots_heen_zh_en}
\end{figure*}

\paragraph{\Comet scores for original vs corrected translations.} 
In Figure~\ref{fig:scatter_plot} (in \S\ref{sec:refining_translations}), we show how \xTower behaves depending on the quality of the original translation for \textsc{en-de} samples. Now, in Figure~\ref{fig:scatter_plots_heen_zh_en} we show plots for \textsc{he-en} and \textsc{zh-en}. Overall, we observe that the same trend remains: \xTower is particularly helpful for cases where the original translation obtains weak-moderate \Comet scores (from 0 to 80\%).

\begin{table*}[t]
    \small
    \centering
    \setlength{\tabcolsep}{3.0pt}
    \begin{tabular}{l@{\quad} ccccc c@{\ \ } ccccc c@{\ \ } ccccc}
        \toprule
        & \multicolumn{5}{c}{\sc en-de} & & 
        \multicolumn{5}{c}{\sc he-en} & & 
        \multicolumn{5}{c}{\sc zh-en} \\ 
        \cmidrule{2-6} \cmidrule{8-12} \cmidrule{14-18} 
        
        \sc Model & 
        chrF & bleu & bleurt & comet & ckiwi & &
        chrF & bleu & bleurt & comet & ckiwi & &
        chrF & bleu & bleurt & comet & ckiwi \\
        \midrule

        Original MT & 64.8 & 39.0 & 48.4 & 78.4 & 75.5 & & 56.5 & 33.5 & 59.8 & 77.5 & 75.5 & & 49.6 & 23.8 & 55.2 & 78.0 & 76.7 \\
        
        \cdashlinelr{1-18}
        \multicolumn{16}{l}{\textcolor{gray}{\textit{Translation-only LLMs:}}} 
        \\
        
        Mixtral 8x7B & 61.5 & 32.4 & 46.4 & 80.4 & 76.6 & & 50.9 & 24.5 & 53.9 & 71.6 & 69.3 & & 46.5 & 17.0 & 53.5 & 77.7 & 77.3 \\
        GPT 3.5T & 68.2 & 41.9 & 51.3 & 82.7 & 78.6 & & 64.5 & 43.9 & 65.5 & 80.9 & 77.8 & & 50.2 & 22.0 & 57.1 & 79.9 & 79.2 \\
        \sc TowerInst 13B & 66.3 & 40.1 & 50.0 & 82.2 & 78.7 & & 45.9 & 22.6 & 50.7 & 68.7 & 66.5 & & 48.3 & 21.6 & 56.5 & 79.1 & 78.4 \\
        \xTower 13B & 66.5 & 40.0 & 50.5 & 82.2 & 78.6 & & 45.8 & 22.2 & 50.6 & 69.4 & 67.3 & & 48.7 & 21.6 & 56.8 & 79.5 & 78.5 \\

        \midrule

        \multicolumn{18}{c}{\textit{Referenceless}}
        \\
        
        \multicolumn{16}{l}{\textcolor{gray}{\textit{With predicted error spans:}}} 
        \\

        Mixtral 8x7B & 29.9 & 10.4 & 42.9 & 64.9 & 58.7 & & 53.5 & 31.6 & 58.1 & 76.4 & 73.2 & & 41.5 & 18.1 & 51.2 & 74.4 & 73.4 \\
        GPT 3.5T & 62.8 & 37.5 & 53.4 & 81.6 & 77.5 & & 60.0 & 38.2 & 63.9 & 80.9 & 77.9 & & 48.6 & 22.1 & 56.2 & 79.1 & 77.9 \\
        \xTower 13B & 59.5 & 34.1 & 52.7 & 81.3 & 77.0 & & 57.1 & 34.5 & 60.9 & 78.5 & 75.6 & & 48.5 & 20.8 & 56.0 & 79.0 & 78.4 \\
        + Hybrid & 64.8 & 38.4 & 52.4 & 82.2 & 80.1 & & 59.9 & 37.4 & 62.4 & 80.0 & 78.7 & & 51.4 & 24.1 & 55.4 & 79.1 & 78.8 \\
        
        \cdashlinelr{1-18}
        \multicolumn{16}{l}{\textcolor{gray}{\textit{With human-annotated error spans:}}} 
        \\

        Mixtral 8x7B & 37.7 & 16.4 & 42.1 & 66.8 & 61.3 & & 54.0 & 30.7 & 57.7 & 76.0 & 73.1 & & 43.3 & 19.4 & 52.8 & 75.7 & 74.1 \\
        GPT 3.5T & 63.1 & 37.6 & 50.2 & 80.6 & 76.5 & & 58.6 & 36.1 & 62.6 & 80.0 & 77.4 & & 48.8 & 22.3 & 56.6 & 79.2 & 77.9 \\
        \xTower 13B & 61.3 & 35.3 & 50.2 & 81.3 & 77.3 & & 56.3 & 33.3 & 60.0 & 77.7 & 75.0 & & 49.2 & 21.3 & 56.4 & 79.4 & 78.6 \\
        + Hybrid & 64.7 & 38.4 & 52.7 & 82.5 & 79.9 & & 60.3 & 38.2 & 63.6 & 80.8 & 79.4 & & 51.7 & 24.6 & 56.2 & 79.7 & 79.2 \\

        \midrule

        \multicolumn{18}{c}{\textit{Reference-based}}
        \\
        
        \multicolumn{16}{l}{\textcolor{gray}{\textit{With predicted error spans:}}} 
        \\

        Mixtral 8x7B & 36.9 & 16.1 & 49.4 & 70.0 & 62.8 & & 71.2 & 54.2 & 74.1 & 85.6 & 77.2 & & 53.2 & 31.0 & 62.4 & 80.6 & 75.2 \\
        GPT 3.5T & 74.1 & 55.4 & 63.3 & 85.2 & 78.6 & & 79.1 & 65.3 & 80.2 & 88.8 & 79.3 & & 58.8 & 36.6 & 66.5 & 83.3 & 77.7 \\
        \xTower 13B & 70.0 & 50.8 & 62.9 & 84.6 & 77.7 & & 81.0 & 66.2 & 80.5 & 89.0 & 79.4 & & 60.1 & 35.9 & 66.8 & 83.7 & 78.3 \\
        + Hybrid & 73.4 & 52.7 & 62.4 & 85.8 & 79.4 & & 82.3 & 69.5 & 80.2 & 88.4 & 79.5 & & 63.6 & 39.8 & 66.5 & 83.8 & 78.1 \\
        
        \cdashlinelr{1-18}
        \multicolumn{16}{l}{\textcolor{gray}{\textit{With human-annotated error spans:}}} 
        \\

        Mixtral 8x7B & 40.3 & 18.7 & 46.3 & 69.9 & 63.5 & & 69.0 & 50.4 & 72.1 & 84.5 & 76.4 & & 51.6 & 29.1 & 60.7 & 80.0 & 75.3 \\
        GPT 3.5T & 71.6 & 50.9 & 58.5 & 83.7 & 77.9 & & 75.9 & 60.8 & 77.7 & 87.6 & 78.7 & & 57.7 & 35.0 & 65.3 & 82.9 & 77.9 \\
        \xTower 13B & 70.6 & 50.7 & 59.1 & 84.1 & 77.9 & & 78.9 & 62.5 & 78.8 & 88.2 & 78.7 & & 59.8 & 35.6 & 66.5 & 83.6 & 78.5 \\
        + Hybrid & 73.5 & 52.5 & 61.7 & 86.0 & 79.7 & & 80.6 & 67.1 & 79.6 & 88.6 & 80.1 & & 63.3 & 39.7 & 67.1 & 84.3 & 78.5 \\

        \bottomrule 
    \end{tabular}
    \caption{Full results for translation correction experiments in terms of lexical and neural metrics.}
    \label{tab:full_results_refinement_appendix}
\end{table*}

\section{Computational Details}

All experiments involving \xTower and Mixtral 8x7B were carried on Nvidia RTX A6000 GPUS with 48GB VRAM. 
For GPT 4 and GPT 3.5T, we used the official API from OpenAI.
We used VLLM\footnote{\url{https://github.com/vllm-project/vllm}} for efficient generation.

\section{AI Assistants}
We have used Github Copilot\footnote{\url{https://github.com/features/copilot}} during code development, and ChatGPT\footnote{\url{https://chat.openai.com/}} during paper writing for grammar correction.

\end{document}